\documentclass[fleqn,10pt]{wlscirep}
\usepackage[utf8]{inputenc}
\usepackage[T1]{fontenc}

\usepackage{graphicx}
\usepackage{caption}
\usepackage{subcaption}

\usepackage{hyperref}
\usepackage{url}

\usepackage{booktabs}

\usepackage[utf8]{inputenc}
\usepackage{newunicodechar}
\newunicodechar{≥}{\geq}
\usepackage{tabularx}
\usepackage{booktabs}
\usepackage{enumitem}
\usepackage[colorinlistoftodos,textsize=small]{todonotes}
\usepackage{tabularx}
\usepackage{booktabs}
\usepackage{colortbl}
\usepackage{xcolor}
\usepackage{algorithm}
\usepackage{blindtext}
\usepackage{authblk}
\usepackage{hyperref}
\usepackage{algorithmic}
\usepackage[font=small,labelfont=bf]{caption}
\title{ 

Topological texture analysis of microscopy images of dynamic casein gelation and its relation to rheological properties
    
}

\author[1,+]{Zahra Tabatabaei}
\author[2,+]{Diana Soto-Aguilar}
\author[3]{José C. Bonilla}
\author[2]{Mathias P. Clausen}
\author[1*]{Jon Sporring}
\affil[1]{Department of Computer Science, University of Copenhagen, Denmark.}
\affil[2]{Department of Green Technology, University of Southern Denmark, Denmark.}
\affil[3]{Department of Food Science, University of Copenhagen, Denmark.}

\affil[*]{sporring@di.ku.dk}

\affil[+]{these authors contributed equally to this work}

\begin{abstract}
We propose a novel computational toolbox that integrates Topological Data Analysis (TDA), Differential Box Counting (DBC), Multifractal Partition (MFP), and Local Binary Patterns (LBP), applied to time-lapse super-resolution STED microscopy images of sodium caseinate gelation induced by glucono-$\delta$-lactone (GDL) at 30 °C and 40 °C and two GDL concentrations (1.8\% and 3.5\% w/v). 

TDA tracked topological loops, closed ring-like structures reflecting protein network interconnectivity, via max-Betti-1 curves, which revealed a lag phase of dispersed aggregates, a sharp decay coinciding with network percolation and the rheologically observed sol-gel transition, and a post-gelation increase corresponding to network rearrangements. These topological transitions were corroborated by DBC and MFP as these methods were able to resolve changes in structural complexity and spatial heterogeneity. The toolbox was validated on simulated fractal images prior to experimental application. Together, these descriptors provided sensitivity to subtle microstructural transitions that bulk rheology captured as averaged bulk mechanical responses. This integrated approach provides a robust quantitative tool for characterizing complex microstructure in food and material science with evolving microstructural dynamics . Code is available at \href{https://github.com/Zahratabatabaei/Delifood_CV_paper.git}{this GitHub repository}.

\end{abstract}

\begin{document}

\flushbottom
\maketitle

\thispagestyle{empty}


\section*{Introduction}

Acidification of milk proteins underlies the formation of protein gels in a range of fermented and acid-set dairy products, where a controlled pH decrease drives aggregation into a gel network. For research, the use of sodium caseinate (NaCas) and glucono-$\delta$-lactone (GDL) has been extensively used as a model system as these provide a controlled gradual decrease in pH and a reproducible gelation behavior. Specifically, the acidification of NaCas via GDL causes a decrease of protein net charge near the isoelectric point, which leads to electrostatic repulsion and non-covalent aggregation resulting in an interconnected network, a gel \cite{raak2025strain}. The functional properties of these acid-induced casein gels that are relevant for the food industry, including water-holding capacity, syneresis susceptibility, and rheological behavior, are determined by the spatial organization of the aggregated proteins within the network, which is why gel microstructure has been characterized in terms of its porosity, degree of connectivity, and fractal-like organization \cite{hannss2020acid, pugnaloni2005microstructure}. However, such characterization often remains descriptive due to a lack of specialized tools for quantitative image analysis \cite{saalbrink2025quantifying}, thus, developing more quantification techniques is essential to mathematically link microstructural observations with their macroscopic behavior.

Casein gelation is also inherently dynamic, characterized by an initial formation of the gel network and subsequent bond strengthening and/or local rearrangements \cite{braga2006effect}. However, this contrasts with the static microscopic view often used for their analysis. While the two stages can be identified with bulk techniques like rheology and diffusing wave spectroscopy \cite{braga2006effect, raak2016potential}, they are indirect measurements that average signals over the entire sample volume, so they do not effectively capture subtle structural changes or overlying spatial arrangements. Thus, high-resolution direct imaging is required to resolve the structural changes determining these bulk properties. Specifically, super-resolution imaging techniques can be used to better understand structure-function relationships in food matrices \cite{bonilla2025advances}.

In this sense, Stimulated Emission Depletion (STED) microscopy is relevant as it is used to bypass the diffraction limit to enhance resolution with a selective fluorescence confinement via a donut-shaped depletion laser \cite{bonilla2025advances}. For example, STED microscopy has previously been used to study egg protein networks in cooked eggs with different rheological textural attributes \cite{bonilla2022super}, where the enhanced resolution allowed an enhanced quantitative characterization of protein particles that correlated with the macroscopic viscoelastic properties. Other studies have extracted microstructural features in dairy gels with specialized image analysis techniques that enabled a detailed quantification of different processing conditions \cite{glover2019super}, of water dynamics in gels with added pectins \cite{mishra2025quantifying}, and of casein and whey proteins interactions in mixed systems \cite{li2023discriminating}. These studies relied on pixel intensity metrics to characterize such systems, however, when monitoring dynamic casein networks, this approach may be limited by potential bleaching from prolonged imaging. Due to the dynamic behavior, the evolving three-dimensional structures can drift out of the focal plane, which also compromise pixel intensity-based methods by introducing quantification artifacts. To overcome this, we propose shifting the analytical focus from intensity statistics to the topological connectivity of the network, a feature that is insensitive to these specific imaging challenges. 

Topological analysis is a standard tool in computational soft matter physics, but its application to experimental food systems remains mostly unexplored. In this paper, we demonstrated that topology can be used to quantify their mechanical behavior with features like loop formation, since this feature correlated with the gel response under deformation~\cite{bouzid2018network}. This study provides a theoretical basis for applying topological quantification to caseinate gelation. We hypothesize that combining topological frameworks with super-resolution STED microscopy will yield quantitative topological insights into the acid gelation mechanism.

We are presenting a new data-driven toolbox for understanding the dynamic changes of acid casein gels. We propose a global texture analysis technique to extract the topological and textural properties of the dynamic protein aggregation of sodium caseinate as a model system using STED microscopy.
The main contribution of this paper is as follows:

\begin{itemize}
    \item Introducing a novel analytical framework integrating Topological Data Analysis (TDA), Differential Box Counting (DBC), Multifractal Partition (MFP), and Local Binary Patterns (LBP) to characterize the dynamic microstructural evolution of sodium caseinate gels.

    \item Acquiring high-resolution STED microscopy time series (145 frames, one every 25s, at 30\,nm pixel size) during acid-induced gelation at 30\,°C and 40\,°C to map protein network formation under varying pH and acidification rates.

    \item Demonstrating the first application of TDA to dynamic STED microscopy data, revealing topological transitions to perform an accurate and invariant analysis of topological features of texture beyond intensity-based properties.

    \item Quantifying and correlating topological (TDA), fractal (DBC, MFP), and textural (LBP) descriptors with a macroscopic rheological property, storage modulus (G').

    \item Validating the proposed methodology on both simulated and experimental data, showing high sensitivity to early aggregation events and subtle post-gelation network rearrangements.
\end{itemize}

\subsection*{Related works}

Recent studies on food and Artificial Intelligence (AI) illustrate the recent advancements of food science by using AI-driven tools to analyze the food patterns~\cite{kharbachAIPoweredAdvancesData2025}. Gao et al.~\cite{gaoMachineLearningdrivenInnovations2025} provided a comprehensive review of data-driven modeling in food processing. They summarize ML applications in food quality detection, drying, and fermentation, noting ML’s strong nonlinear modeling capabilities for multi-factor processes. Siddique et al.~\cite{BigDataAnalytics} examine AI/ML solutions for key challenges in food safety, quality, and supply chain. The review consolidates how AI can enhance food production and monitoring, bridging gaps due to previously fragmented studies. While these studies emphasize the performance and applications of AI in food science, other research has investigated structural and mechanical behavior in food systems using AI-driven analytical methods. 

Auty et al.~\cite{autyDynamicConfocalScanning1999} developed dynamic confocal scanning laser microscopy methods to observe milk protein gelation and cheese melting in real time. By time-lapse confocal imaging, they visualized casein network formation and fat globule movement during rennet-induced milk coagulation and cheese heating. Quevedo et al.~\cite{pedreschiCharacterizationFoodSurfaces2000} found that fractal functions effectively model the rough, granular surfaces typical of natural foods. Andoyo et al.~\cite{andoyoFractalDimensionAnalysis2018} reviewed the application of fractal dimension analysis to whey protein-based gels, comparing image-derived textural features, where texture denotes the spatial distribution of pixel intensities in microscopic images rather than a mechanical material property, with rheological measurements. The authors confirmed that both methods consistently describe the aggregation regime of whey protein gels despite the respective limitations. Xu et al.~\cite{FractalAnalysisAggregates} performed image-based fractal analysis on protein gel aggregates. They also found that fractal analysis is a powerful tool to quantify the irregular geometries of protein aggregates in heat-induced gels. Dubey et al.~\cite{lahiriSurfaceCharacterizationProteins2008} claimed that the multifractal roughness descriptors can be used to compare experimental protein aggregates with simulation results. Senga Yukako et al.~\cite{sengaInSolutionMicroscopicImaging2019} demonstrated that the fractal dimension is a highly useful parameter for characterizing protein aggregate morphology, beyond traditional size measures like radius or mass. Overall, results of protein gels showed that a higher fractal dimension indicated more compact network with densely branched aggregates.

Microscopic images are inherently subject to variability from fluctuating illumination, scale, rotation, and sample deformation \cite{xu2025multi}. Since fractal-based descriptors alone may be insufficient to capture this variability robustly, complementary image-based textural quantification methods should be incorporated for a more comprehensive characterization. Local Binary Patterns (LBP), due to its unique properties and strength, such as strong local texture discrimination, computational efficiency, and 
robustness to monotonic gray-scale changes, represents a suitable technique for extracting such image-based texture features without being compromised by illumination-based artifacts. Ojala et al.~\cite{ojala2001generalized} introduced LBP in 2001, which is a histogram over the occurrences of all binary patterns in the image. LBP is computationally low-cost and efficient in its discrimination power to represent image-based texture features~\cite{ojala2001generalized}, and it is inherently invariant to illumination and rotation~\cite{xu2025multi}. So, LBP and its variants have been successfully applied to various image-texture analysis tasks~\cite{hadid2015gender, huang2011local, zhang2021bioinspired}. For instance, LBP in~\cite{wang2019classification} is used to retrieve
texture-based features and recognize patterns from the plantar pressure optical imaging dataset. In the study by Banerjee et al. ~\cite{banerjee2018local}, to capture the rich image-texture information carried by neighboring pixels, it was proposed a local neighborhood intensity pattern (LNIP) for image retrieval. However, these methodologies do not ultimately address the lack of domain-specific priors, such as the topological prior that can capture a better connectivity of the image-based texture~\cite{banerjee2020semantic}.

Recent advances have demonstrated the potential of Topological Data Analysis (TDA) for understanding soft material structure, particularly in simulation environments. Shah et al.~\cite{hussain2025topological} applied persistent homology to dynamic systems and showed that Betti curves and persistence descriptors can capture regime shifts and temporal evolution in simulated nonlinear systems, demonstrating the strength of TDA for time-dependent phenomena. Smith et al.~\cite{smith2024topological} used the Euler characteristic and filtration-based topological summaries to characterize the hierarchical structure of particulate gels generated by molecular simulations, revealing how material parameters such as quench rate and volume fraction influence the emergence of loops, cavities, and voids across multiple length scales. TDA and persistence diagrams are used in~\cite{minamitani2025persistent} to quantitatively capture and differentiate the geometric and structural features of chromatin organization, enabling classification of cell types and conditions. The power of persistent homology was leveraged to analyze high-dimensional data in~\cite{zabaleta2025unveiling} to identify synchronization transitions by examining local connectivity structures, serving as a benchmark to characterize the collective behavior of systems modeled by oscillator networks.

Although these studies highlight the power of topology-based descriptors for soft matter characterization, their analyses remain purely computational and have not yet been validated on experimental microscopy data. To date, no study has examined whether topological signatures derived from real protein aggregation processes align with physical gelation behavior or correlate with measurable rheological properties. This gap motivates the present work, where we experimentally apply cubical complex-based TDA to super-resolution microscopy images of acid-induced protein gels and evaluate how topological evolution relates to the underlying aggregation process.

\section*{Material and Methodology }\label{method}



\subsection*{Simulated data set}

To evaluate the behavior of the proposed framework under controlled conditions, we first analyzed simulated data with parameters and visual characteristics similar to the experimental images. To this end, 30 synthetic images were generated using a fractal generator (\href{https://icefractal.com/fractalator/}{Fractalator}), with the roughness parameter systematically increased from 0.7 to 1.0, as shown in Figure~\ref{fig:simulated}. In this data set, increasing roughness corresponds to progressively more irregular, fragmented, and heterogeneous spatial patterns, allowing controlled variation of structural complexity across the dataset. These images exhibit both fractal and multifractal-like characteristics, making them suitable for evaluating descriptors such as DBC and MFP. The number of generated images was chosen to provide sufficient sampling of the structural transition while maintaining computational efficiency. These values were empirically selected to mimic patterns observed in the experimental dataset.

Although these simulated images do not fully reproduce the complexity of real protein networks, they provide a controlled environment for assessing the sensitivity of the proposed analysis framework. After carefully implementing the techniques and analysing the results, the framework was subsequently applied to the experimental STED images obtained from NaCas.

\subsection*{Data set aquisition}\label{real}

Sodium caseinate powder (CAS 9005-46-3; ${\geq}$$92\%$ protein on a dry basis; Thermo Scientific Chemicals, Waltham, MA, USA) was dissolved in demineralised water at a concentration of $2.6\%$ w/v, and the dispersion was stirred with a magnetic stirrer overnight at room temperature (22°C). 

Fresh sodium caseinate solution ($2.6\%$ w/v) was fluorescently labelled for STED imaging by adding Abberior STAR RED-NHS (26.2 mM) (Sigma-Aldrich, Denmark) at a dye:sample ratio of 0.3:1000. Immediately after staining, glucono-$\delta$-lactone (GDL) was added at a concentration of either 1.8 or $3.5\%$ w/v, and mixed vigorously with a vortex for 1 min before imaging. After, 150 $\mu$L were transferred to the \textit{smart substrate} (18.0 x 18.0 x 0.17 mm$^3$ (\#1.5H)) of a VAHEAT heating stage, which was then heated at 30°C and 40°C before the recording started. A coverslide was used to cover the sample to reduce evaporation. The imaging was initiated 3 min after the initial contact of the sample with the GDL. STED microscopy was performed using an Abberior Facility Line STED system (Abberior Instruments GmbH, Germany) with an objective A UPLXAPO 60X/1.42 NA oil immersion (Olympus). The pinhole was set to 1 AU. Images were acquired with single-channel excitation (640 nm) and 775 nm depletion laser. Fluorescence emission was detected within a spectral window of 650–755 nm under gated detection (750 ps / 8 ns). Imaging was conducted using 0.5\% excitation laser power and 3\% STED laser power with three STED scans per frame. The pixel dwell time was set to 15 µs, and the imaging depth was maintained at 5$\mu$m to ensure high-resolution acquisition while preserving structural integrity.

In this study, a $30~\mu\text{m} \times 30~\mu\text{m}$ field of view was captured in the $xy$-plane using an optimized pixel size of 30~nm, resulting in images of $1000 \times 1000$ pixels. A total of 145 frames were acquired, due to the instrument settings each was taken every 25~s, yielding one continuous hour of imaging.

\subsection*{Rheological measurements}
Oscillatory rheological measurements were done to correlate the quantification of the network microscopic formation in the dynamic set with their mechanical properties, i.e., the transition from a liquid-like solution to a viscoelastic gel due to the acidification with glucono-$\delta$-lactone (GDL). Similarly to the sample preparation for microscopy imaging, fresh sodium caseinate solution (\% w/v) was mixed with GDL at either 1.8 or 3.5\% w/v and mixed vigorously for 1 min. After, 12 mL of the solution was transferred to a double-gap concentric cylinder geometry (inner radius 15.1 mm, outer radius 18.5 mm, bob length 53 mm) on a stress-controlled rheometer (HR20 TA Instruments). A Peltier device maintained the specific temperature, either 30°C or 40°C. The network formation was monitored at a strain amplitude of ($\gamma$) of 0.1\% and a frequency of 1 Hz. The measurement started 3 min after the initial contact of GDL with the sample to match the imaging protocol. All measurements were performed in duplicate. Results are described as the storage modulus (G'), which represents the elastic response of the material and is used as an indicator of the gel stiffness.

\subsection*{Topological Data Analysis (TDA)}
\begin{figure}[H]

\centering
\begin{subfigure}{0.65\textwidth}
    \centering
    \includegraphics[width=\linewidth]{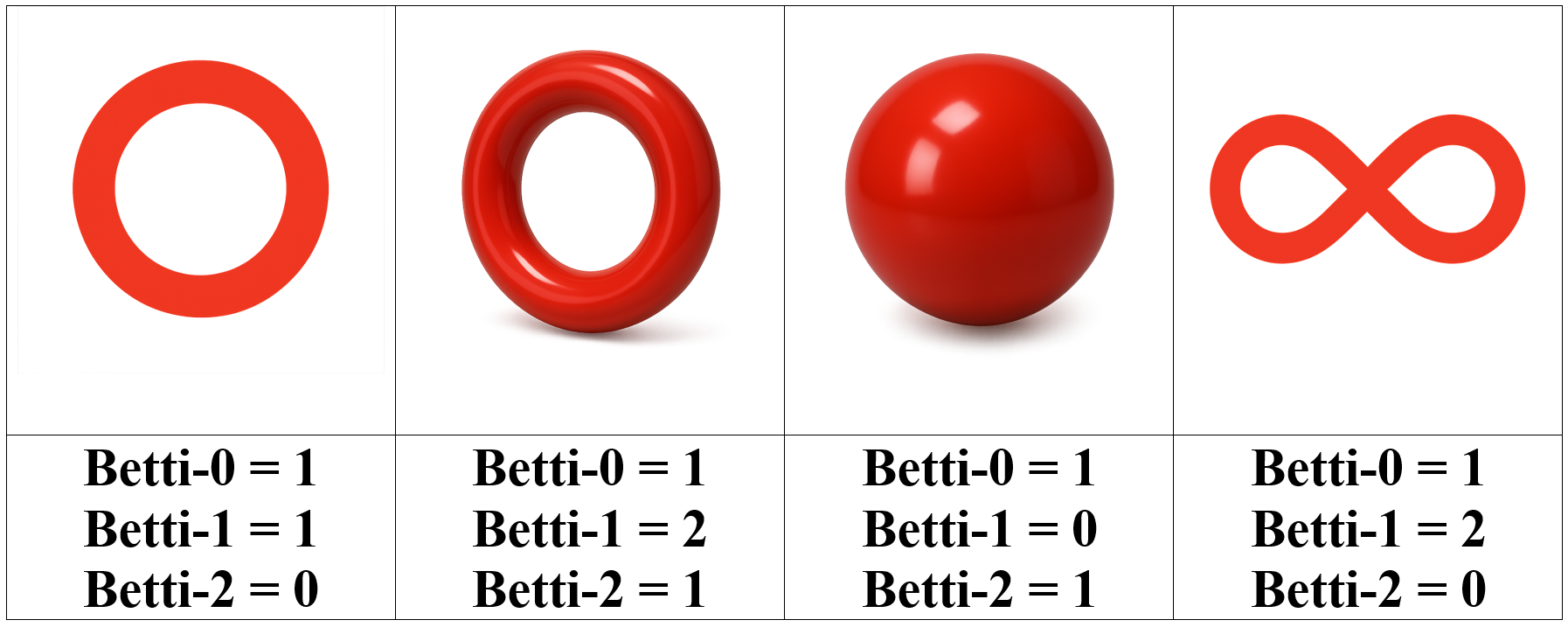}
    \caption{Simple representation of Betti values. The first and the last examples are in 2D, and the two examples (a torus and a sphere) in the middle are in 3D.}
    \label{fig:bettis}
\end{subfigure}
\vspace{1em}
\begin{subfigure}{0.75\textwidth}
    \centering
    \includegraphics[width=0.90\linewidth]{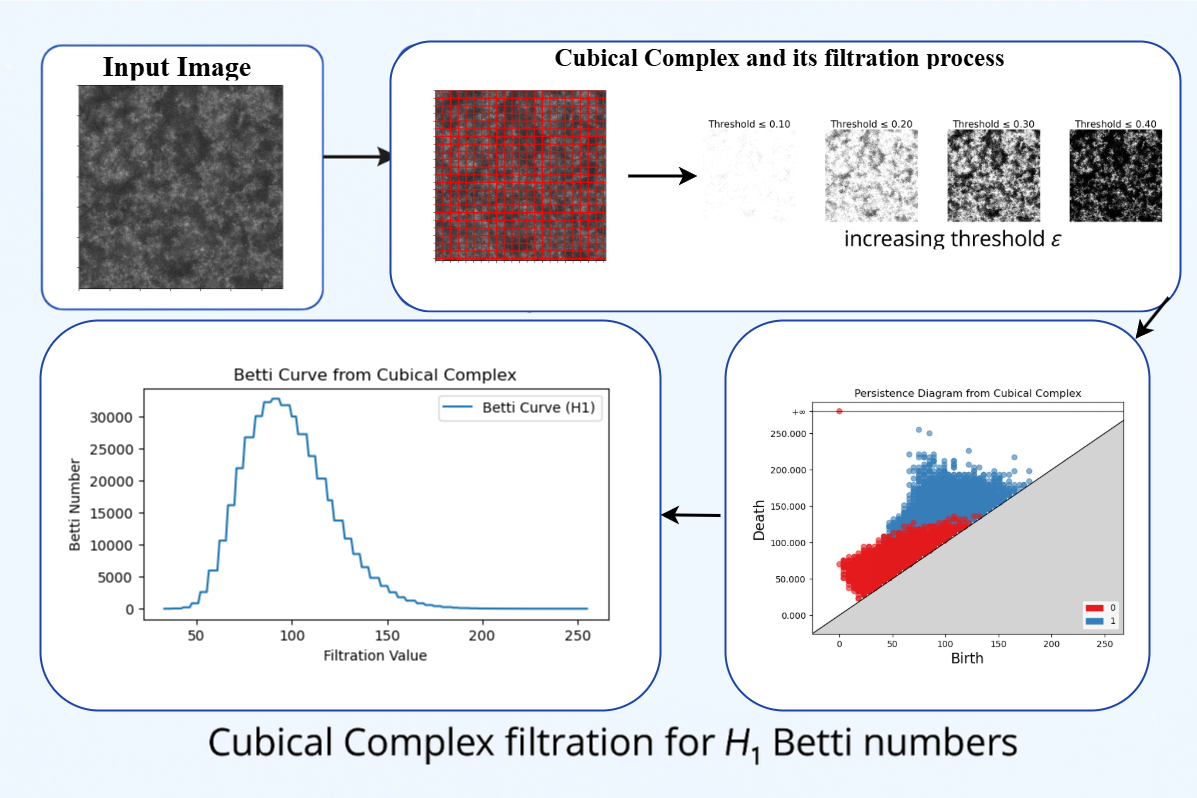}
    \caption{Overview of the cubical complex–based topological analysis pipeline. The input image is converted into a cubical complex and filtered across increasing threshold values. The resulting persistence diagram and Betti curve characterize the evolution of $H_1$ topological features (loops) across the filtration.}
    \label{fig:Cubical_complex}
\end{subfigure}

\caption{Topological background and cubical complex pipeline used for persistence-based analysis.}
\label{fig:topology_overview}

\end{figure}

Topological Data Analysis (TDA) is an promissing approach for food structure characterization that leverages the underlying topology of images within the food domain. TDA focuses on capturing the global structure of an image, such as loops and holes that remain invariant to specific geometric measurements~\cite{tabatabaei2025thir}. This framework characterizes images through their fundamental topological elements: 0D features (Betti-0: $\beta_0$), representing connected components; 1D features (Betti-1: $\beta_1$), corresponding to loops; and 2D features (Betti-2: $\beta_2$), denoting voids or enclosed surfaces. Figure~\ref{fig:bettis} shows four simple samples, including a hollow, a torus, a sphere, and an infinity symbol. In the second column of the figure, the Betti values for each shape have been reported. For instance, a torus is one connected component, which yields $\beta_0 = 1$. In addition to $\beta_0$, it has two loops, which mathematically equals $\beta_1 = 2$. Also, since it is a 3D shape, the $\beta_2 = 1$. Characterizing the images with TDA provides a comprehensive representation of an image’s structural and textural properties. Importantly, TDA serves as a powerful shape-based feature extractor that is inherently invariant to rotation and scaling.

TDA has several computational techniques for extracting topological features, including Ripser, the Čech complex, and the cubical complex. Among these, the cubical complex is particularly well-suited for image data, as it operates directly on regular pixel grids. In this framework, each unique pixel intensity serves as a threshold value. Consequently, instead of applying a single global threshold, which simplifies the image into a binary form and omits significant structural details, the cubical complex performs thresholding across all unique intensity values present in the image.

As the output, TDA offers several quantitative outputs for image analysis, including persistence homology, Betti curves, and persistence landscapes. In the context of this study, Betti curves were chosen as they provide interpretable and comparative insights for food scientists, particularly when correlated with traditional analytical measures such as rheological curves.

Figure~\ref{fig:Cubical_complex} further demonstrates how the cubical complex processes each input image by overlaying a grid on it. Every unique pixel intensity within this range acts as a threshold value ($th$). As the threshold increases, loops (corresponding to $\beta_1$ features) emerge (are born) and merge (die). The persistence homology diagram illustrates the lifetime of each $\beta_1$ feature across varying thresholds, where the diagonal line represents $Birth = Death$. Points farther from the diagonal indicate more persistent and thus more meaningful topological features~\cite{adler2010persistent}.

From this analysis, a Betti curve can be derived for each frame, typically resembling a bell-shaped distribution. At lower threshold values, few loops are observed; as the threshold increases, the number of loops rises and subsequently decreases as the image darkens. The Max-Betti value for each frame is then identified, representing the highest number of loops observed for all thresholds. Finally, we define the sequence of these maximum values across all frames as the max-Betti-1 curve, which encapsulates the temporal evolution of the topological structure of the system. This curve provides valuable insights into the aggregation behavior of the sample from a topological perspective. The Max-Betti-1 operator is invariant under affine transformation of the intensities, i.e., if the imaging conditions are changed, then the max number will remain essentially the same. This feature of TDA makes it a promising technique to analyze the microscopic images.

As it is mentioned above, TDA captures the topological descriptor of the network. For further investigation of intensity-based variations, fractal analysis was employed using DBC, which quantifies grayscale structural complexity.

\subsection*{Fractal dimension}

Typically, a Fractal Dimension (FD) analysis uses a binarized image, which yields a single value as the output showing the dimensionality~\cite{wu2025multifractal}. In the case of protein aggregation and its complex process (clusters, voids, coarsening), studies have shown that FD alone is not able to properly quantify casein networks as qualitatively different structures give similar FD values, while in other cases, thresholding biases may compromise results~\cite{mellema2000structure, pugnaloni2005microstructure}. Fractal measurements are typically performed on static images of the final gel network, where structural differences are pronounced. As our study explores the dynamic behavior of the proteins, small changes will not be captured due to the mentioned limitations.

In this work, we applied Differential Box Counting (DBC). This method treats the image as a 3D surface (x, y = space, z = intensity) and handles grayscale images without binarization. DBC accounts for both spatial and intensity fluctuations, making it sensitive to subtle changes during protein aggregation. DBC partitions the image into spatial boxes of size \(x \times y\), while the grayscale intensity range is treated as a third dimension \(z\) to quantify local intensity variation. In this method, \( z \) can take values in the range \([0, 255]\) for 8-bit integer images. To analyze the scaling behavior across multiple spatial resolutions, the image was partitioned using logarithmically spaced box sizes, ensuring balanced sampling of both small and large spatial structures.

While DBC operates directly on grayscale intensity surfaces, the Multifractal Partition (MFP) implementation used in this work defines the spatial measure from binary protein-occupancy maps obtained via Otsu thresholding. In contrast to DBC, where thresholding would discard relevant grayscale complexity, the objective of MFP here is to quantify the spatial heterogeneity of protein-rich and protein-poor regions across scales. Therefore, binarization provides an interpretable occupancy measure for distinguishing clusters and voids during aggregation. We note, however, that MFP can also be formulated directly on grayscale images by defining the probability measure from normalized pixel intensities.

MFP describes the distribution of different scaling behaviors within a complex structure~\cite{sporring1999information}. This means that, unlike a monofractal with one uniform fractal dimension, which is characterized by a single singularity exponent, a multifractal object contains many local scaling exponents. Such behavior is particularly relevant for protein aggregation systems, where the formation of clusters, pores, and heterogeneous network structures leads to spatial distributions that vary across multiple length scales. Multifractal analysis is therefore well-suited to capturing the hierarchical organization and heterogeneity of evolving protein networks.

In this study, each image was first converted into a binary map of protein occupancy using Otsu thresholding. Then, the image was divided into non-overlapping boxes across multiple spatial scales to characterize the structure across resolutions. To probe the scaling behavior across multiple spatial resolutions, six box sizes were selected using logarithmic spacing with base 2, and the corresponding normalized scales were defined as $\epsilon = \text{box size}/L$, where $L$ is the image side length. For each scale $\epsilon$, the image is partitioned into non-overlapping boxes of size $\epsilon L$. The probability measure inside each box is computed as:

\begin{equation}
P_i(\epsilon)=\frac{N_i(\epsilon)+\eta}{N_{tot}+M(\epsilon)\eta}
\end{equation}

where $N_i(\epsilon)$ denotes the number of white pixels contained in box $i$ at scale $\epsilon$, $N_{tot}$ represents the total number of white pixels in the image, $\eta$ is a small positive Laplace smoothing constant, and $M(\epsilon)$ is the number of boxes at scale $\epsilon$. The smoothing term ensures that $P_i(\epsilon)>0$ for all boxes and scales, which is required for evaluating negative moment orders $(q<0)$. Thus, varying $\epsilon$ changes the spatial support of the measure, enabling the scaling behavior of the partition function $Z(q,\epsilon)$ to be analyzed across resolutions. This quantity represents the fraction of the total protein occupancy contained within box $i$ at scale $\epsilon$. High values of $P_i(\epsilon)$ correspond to dense protein-rich regions, whereas low values correspond to sparse regions or voids. Based on this scale-dependent probability measure, the partition function in MFP is defined as:

\begin{equation}
Z(q, \epsilon) = \sum_i P_i(\epsilon)^q
\end{equation}

In this function, $\epsilon$ is the box size (scale), and  \( q \) is the moment order. The parameter \(q\) acts as a weighting exponent that emphasizes different regions of the probability distribution. Positive values of $q$ emphasize boxes with large probabilities (dense regions), whereas negative values emphasize boxes with small probabilities (sparse regions or voids)~\cite{li2016generalized}. For $q=1$, the information dimension was estimated separately using the Shannon entropy formulation, since the standard expression for $D(q)$ becomes indeterminate at $q=1$. In this study, we used $q \in [-5, +5]$ because it provides a balanced sensitivity to both dense clusters ($q>0$) and sparse regions or voids ($q<0$)~\cite{pamula2013effects}. Larger absolute values of $q$ tend to excessively amplify rare events and small statistical fluctuations, which may introduce numerical instability and reduce the robustness of the estimated scaling behavior.

The scaling law that defines the MFP is given by:

\begin{equation}
Z(q, \epsilon) \propto \epsilon^{(q - 1) D(q)}
\label{eq:partition_scaling}
\end{equation}

where $D(q)$ is the generalized (Rényi) dimension corresponding to the order $q$. In practice, $D(q)$ is obtained from the scaling behavior of the partition function across different box sizes. For each value of $q$, the partition function $Z(q,\epsilon)$ is computed for multiple box sizes $\epsilon$, and the relationship between $\log Z(q,\epsilon)$ and $\log \epsilon$ is analyzed. The slope of this log–log relationship corresponds to $(q-1)D(q)$, from which the generalized dimension $D(q)$ is estimated. A nearly constant $D(q)$ curve indicates a monofractal structure with uniform scaling behavior, whereas a decreasing $D(q)$ with increasing $q$ reflects multifractal behavior and increasing structural heterogeneity.

Evidently, different values of $q$ correspond to fractal dimensions with different meanings~\cite{li2016generalized}; for example, $D(0)$ is the capacity dimension, $D(1)$ is the information dimension, and $D(2)$ is the correlation dimension. Therefore, the generalized dimension spectrum $D(q)$ provides a multi-scale characterization of the measure distribution beyond a single fractal dimension. The generalized dimension spectrum, obtained by analyzing $D(q)$ as a function of $q$, is often used to characterize the multifractal properties of a fractal object~\cite{li2016generalized}. Unlike a single fractal dimension, the spectrum captures how scaling behavior varies across different regions of the measure distribution.

To better characterize the heterogeneity of the protein patterns, we computed the difference between the generalized dimensions evaluated at the minimum and maximum moment orders. Because large positive values of $q$ emphasize dense protein clusters while negative values emphasize sparse or void regions, the difference between $D(q_{\min})$ and $D(q_{\max})$ reflects the imbalance between these two structural regimes~\cite{pamula2013effects}. The magnitude of $\Delta D$ represents the degree of asymmetry in the measure distribution of a fractal object and thus reflects the prominence of its multifractal characteristics.

\begin{equation}
\Delta D = D(q_{\min}) - D(q_{\max})
\label{eq:delta}
\end{equation}

Small values of $\Delta$ indicate more uniform scaling behavior, whereas larger values indicate stronger multifractality and greater structural heterogeneity. Therefore, $\Delta$ quantifies the coexistence of dense protein clusters and sparse regions within a single image.

\subsection*{Local Binary Pattern (LBP)}
While DBC and MFP analyze global and multi-scale structural features, LBP is applied to capture complementary information on fine-scale spatial organization. LBP and its variants are the tools for texture classification, offering computational efficiency and robustness to illumination changes with effective results in characterizing the local structure~\cite{ojala2001generalized, gul2025effective}. LBP provides a unified description including both statistical and structural characteristics of a texture~\cite{zhou2008novel}. The term "local pattern" in LBP is defined as the neighborhood of the central pixel based on a selected radius value~\cite{zhou2008novel}.

One of the most important properties of the LBP technique that makes it suitable for microscopic images is its robustness to monotonic gray-scale changes caused, for example, by slow illumination variations during capturing images~\cite{pietikainen2010local}. The generic LBP descriptor can be defined as follows~\cite{pietikainen2011local}, 

\begin{equation}
\text{LBP}_{P,R}(x_c, y_c) = \sum_{p=0}^{P-1} s(g_p - g_c)\, 2^{p}
\label{eq:lbp}
\end{equation}

Here, the $s$ denotes the thresholding function~\cite{pietikainen2011local}. Where,

\begin{equation}
s(z) =
\begin{cases}
1, & \text{if } z \ge 0, \\
0, & \text{if } z < 0
\end{cases}
\label{eq:thresholding}
\end{equation}

\begin{itemize}
    \item $(x_c, y_c)$ is the \textbf{center pixel},
    \item $g_c$ is the \textbf{gray value of the center pixel},
    \item $g_p$ is the \textbf{gray value of the $p$-th neighbor} at radius ($R$),
    \item $P$ is the \textbf{number of sampling points} around the center.
\end{itemize}

Radius (R) and the number of sampling points (P) allow the algorithm to capture texture at different scales and resolutions. P specifies the number of neighbors sampled around the center pixel. More points allow for a more detailed encoding of the local structure. R controls the spatial resolution, and a larger radius value allows the method to capture larger-scale texture features. Sampling a few points at a large R can miss critical local details~\cite{xu2025multi}. So, choosing the correct values for these parameters is critical. The generic LBP operator can be applied to any neighborhood size; however, it is most commonly defined for a \(3\times 3\) neighborhood (\(P=8,R=1\))~\cite{gul2025effective}, resulting in an 8-bit binary number (ranging from 0 to 255) for each pixel.


Then, the binary values for all neighboring pixels are combined to obtain a binary code for the central pixel. This code is obtained by cyclically shifting each binary pattern to get the minimum value~\cite{ojala2001generalized}. After processing the entire image, a histogram of these decimal values is created to serve as the feature vectors. These histograms illustrate how often each pattern occurs. It is noteworthy to mention that while computing the histogram of the LBP results using the $histogram$ function from the NumPy library, the $density $parameter can be set to $True$ or $False$. $Density = False$ is the default mode, which measures the frequency of a pattern occurring. This cannot be used in the case of image comparison because it fails if the number of pixels varies between frames. $Density = True$ measures the probability of a pattern occurring, which quantifies structural complexity. This can work even if the image size of the shape changes. In this case, the histogram no longer counts the number of pixels per bin; it calculates the probability density function. So, the area under the histogram is normalized to sum to 1.
If we set $Density = False$, if we compare to different frames and one has slightly more valid pixels, the raw count entropy will change even if the texture is identical. In the case of this study, texture is what matters most. So, we set $Density = True$ to measure the distribution of patterns as our features under study, not the patterns themselves. By measuring the std and entropy of these features, comprehensive information about the distribution of the textures and local regions can be obtained.

All these methods were applied to the contrast-stretched images. So that their minimum and maximum intensities are mapped to 0 and 255, respectively. In the following section, for the simulated data set and the longitudinal sodium caseinate data set, we report results from TDA, DBC, MFP, and LBP. It is noteworthy to mention that all experimental parameters were kept consistent across datasets, including settings such as the radius and number of sampling points in the Gabor and LBP computations. This ensures comparability of the results and allows observed differences to be attributed to the data rather than methodological variability. All the codes for this paper can be accessed  \href{https://github.com/Zahratabatabaei/Delifood_CV_paper.git}{in this GitHub link}. In these codes, some parameters (e.g., $q$-range in MFP, box sizes in DBC, and descriptor settings in TDA and LBP) have been chosen guided by commonly used values in the literature and by numerical stability considerations. It is noteworthy to mention that, based on our experiments, moderate variations in these parameters did not qualitatively affect the observed temporal trends or the identification of key structural transitions. In particular, consistent behavior was observed when applying the same parameter settings across different NaCas concentrations, temperature conditions, and simulated datasets. Therefore, the reported results in the following section are considered robust with respect to reasonable parameter choices.

\section*{Results and Discussions}

In this paper, we propose four analysis methods to capture complementary aspects of the evolving protein network structure. These methods focused on different features and characteristics of the NaCas at the microscopic level. As it will be explained in the following subsections, TDA characterizes topological connectivity, DBC quantifies grayscale intensity complexity, MFP describes spatial heterogeneity in terms of cluster–void distributions, and LBP captures local texture variations. Together, these descriptors provide a multi-scale and multi-faceted representation of the aggregation process.

\subsection*{Simulated set results}

\begin{figure}
\begin{center}
\vspace{-1cm}
\centerline{\includegraphics[width=0.65\textwidth]{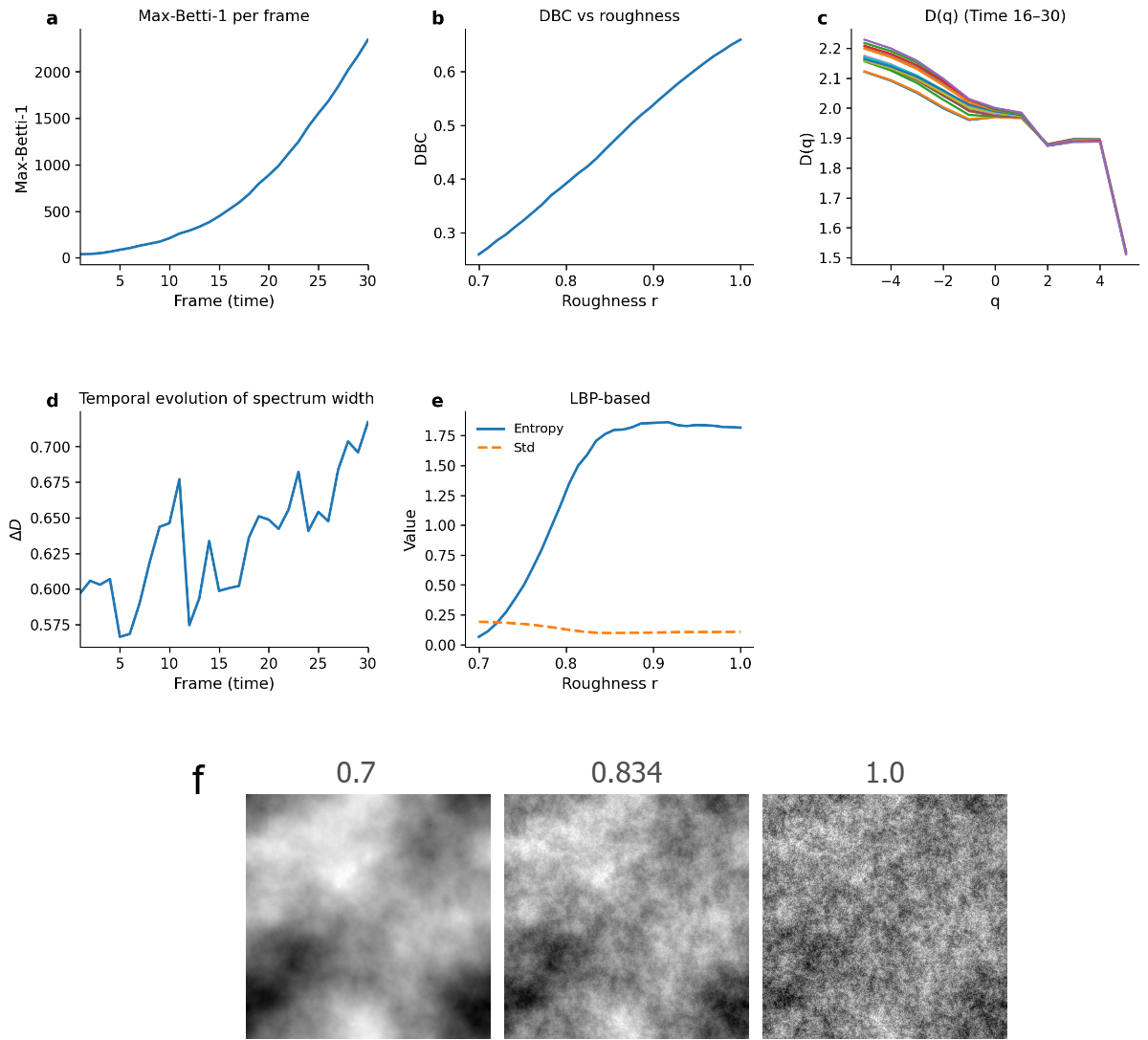}}
\caption{Quantitative descriptors extracted from the simulated fractal dataset. (a) Max-Betti-1 per image, where the frame index corresponds to the 30 simulated images and the y-axis shows the maximum number of loops. (b) DBC results as roughness increases from 0.70 to 1.00. (c) Generalized dimension \(D(q)\) as a function of moment order \(q\) for selected time points (Time 16--30). (d) Temporal evolution of the spectrum width \(\Delta D\), reflecting heterogeneity and dynamic complexity. (e) LBP-based entropy and standard deviation as a function of roughness. (f) three samples of the simulated data with roughness $ = 0.7, 0.834$, and $1.0$.}
\label{fig:simulated_results_all}
\vspace{-1cm}
\end{center} 
\end{figure}

We first applied TDA to the simulated dataset obtained from the fractal generator (\href{https://icefractal.com/fractalator/}{Fractalator}) by increasing roughness over the frames (times). The trend in Figure~\ref{fig:simulated_results_all}a shows that increasing roughness results in a corresponding rise in the max-Betti-1 values. This trend suggests that the max-Betti-1 curve is sensitive to texture complexity and reflects the increasing structural irregularity of the images. The relationship between roughness and max-Betti-1 is presented in Figure~\ref{fig:simulated_results_all}a, where each point represents the max-Betti-1 value extracted from a single simulated image.

We used the DBC approach to estimate the fractal dimension of each frame in order to measure and analyze the structural complexity of the simulated data over the changes in roughness. In this approach, the image at each frame is first partitioned into non-overlapping square grids of multiple scales. For each grid cell, the minimum and maximum gray-level values are used to estimate the number of intensity boxes required to cover the pixel value variation within that region. Summing these intensity box counts across all cells yields $N(s)$ for a given scale $s$. By repeating this process across scales and fitting a linear regression to the relationship between $\log(N(s))$ and $\log(1/s)$, the slope of the fitted line provides an estimate of the fractal dimension. This value reflects how gray-level structure evolves with scale in the simulated textures, allowing us to quantify changes in texture complexity as roughness increases, which reflects the increasing structural complexity of the simulated textures. The results in Figure~\ref{fig:simulated_results_all}b demonstrate a clear monotonic increase in DBC as roughness increases, confirming that the fractal dimension successfully captures the rise in structural complexity across the sequence of textures.  This is similar to what was mentioned in~\cite{sengaInSolutionMicroscopicImaging2019}, which this similarity can support the performance of our method.

It is important to note that this controlled trend does not necessarily replicate the evolution observed in real protein aggregation systems, but rather demonstrates the sensitivity of the DBC descriptor to changes in structural complexity.

Multifractality gives us a compact descriptor of the simulated data to be able to see the changes over frames in an easy and fast way. In Figure~\ref{fig:simulated_results_all}, two of the curves illustrate the MFP over time (\ref{fig:simulated_results_all}c) and the evolution of multifractality width (\ref{fig:simulated_results_all}d). In Figure~\ref{fig:simulated_results_all}c, each line represents the MFP at each frame. The $x-$axis shows which part of the texture we are focusing on based on the $q$ value, while the $y-$axis shows how complex or dimensional those regions are. So, the curve shape explains how intensity heterogeneity changes across scales. In this figure, all the lines start higher on the left (negative $q$), meaning the sparse regions exhibit a higher fractal dimension and greater structural complexity at each frame. This indicates a broad multifractality with many scaling behaviors. Toward the right (positive $q$), we see that the generalized dimension decreases, indicating that the dense regions are more compact and structurally simpler. In the simulated dataset, increasing roughness leads to a broader distribution of scaling behaviors, indicating higher structural heterogeneity. This controlled behavior serves as a benchmark to demonstrate the responsiveness of the multifractal descriptors, rather than a direct representation of protein aggregation dynamics. To summarize the figure and represent the multifractality of the entire dataset relative to surface roughness using a single curve, we calculate the $\Delta$ value. The blue curve in Figure~\ref{fig:simulated_results_all}d shows the evolution of $\Delta$ over time, revealing a clear increasing trend of the multifractality width as the roughness increases. Although the multifractality width $\Delta$ increases over time, indicating a broader range of scaling exponents and higher structural complexity, the generalized dimension curves $D(q)$ appear to converge in the later images. This apparent discrepancy arises because $\Delta$ captures the total spread of scaling behaviors, while the specific values of $D(q)$ reflect the relative dominance of dense and sparse regions. In the simulated dataset, increasing fractality leads to a richer multi-scale structure, resulting in a larger exponent range even as the network becomes more interconnected.

LBP is used in this paper for a rotation-invariant texture analysis. Here we set the LBP method = \textit{uniform} to ignore noisy patterns and only count transitions. This normalizes the patterns by grouping all bit-shifts of the same structural pattern into a single bin. This ensures that the resulting histogram remains consistent even if the sample texture rotates during the gelation process. In this technique, we first applied a Gabor filter (frequencies  $= [0.1, 0.2, 0.3]$, and angles $[0^\circ, 45^\circ, 90^\circ, 135^\circ]$) to the set of images to extract the finer texture details~\cite{gorai2014comparative}. The Gabor filter acts as a band-pass filter to ignore high-frequency noises and only responds to the actual structural edges. As the output of this filter, we will have an image for each combination of frequencies and angles. In this work, the Gabor responses are linearly integrated (summed) into a single 2D spatial map before the LBP is applied. Summing up these outputs is more convincing in the case of this study than concatenating them because it gives a global trend analysis over time. This sums up the results in an integrated map, which gives us a single complexity score for the whole sample. In this integrated map, the aligned parts of the image with the given angles and frequencies are brightened. The filtered images (integrated maps) were then processed using LBP ($n_{\text{points}} = 8 \times \text{radius}$ $=3$). In this paper, 24 sampling points were chosen to see the larger-scale texture to detect global texture variations. Besides, according to recent studies, by enlarging the radius, more global features can be captured, which yields better information about the pixel environment~\cite{lize2019local}. For each resulting LBP map, we computed the corresponding histogram and extracted two statistical descriptors: entropy and standard deviation.

As shown in Figure~\ref{fig:simulated_results_all}e, the entropy increases while the standard deviation decreases as the roughness parameter increases, indicating that higher roughness produces a more uniform and diverse LBP distribution (higher entropy) with reduced variability across histogram bins (lower standard deviation). In Figure~\ref{fig:simulated_results_all}, we exhibit three samples of the simulated data with roughness $ = 0.7, 0.834$, and $1.0$ in order to better view the data and correlate the results with the visual aspects of the data. This is a portion of the entire data set, which is shown in Figure~\ref{fig:simulated}. It is important to emphasize that these experiments on the simulated dataset, which has systematically varied structural complexity through the roughness parameter, provide a controlled framework for better understanding the behaviour of the techniques. As such, the observed monotonic trends primarily reflect the sensitivity of the applied descriptors to increasing heterogeneity and multi-scale structure. In contrast, real protein aggregation involves multiple competing processes, such as cluster formation, network rearrangement, and coarsening, which can lead to non-monotonic descriptor behavior. Therefore, the simulated and experimental results are considered complementary to validate the proposed descriptors under controlled conditions, while in the case of the protein data, they capture the complexity of the gelation dynamics.

\subsection*{Sodium caseinate results}

Results of the simulated data of increasing roughness showed that max-Betti curves can accurately track changes based on topological structure rather than intensity or texture alone. To further demonstrate the use of max-Betti curves, we tracked the gelation of sodium caseinate, and to obtain different final network structures, two temperatures and GDL concentrations were studied. Figures~\ref{fig:FrameBettiph_reps_1.8} and~\ref{fig:FrameBettiph_reps_3.5} show the max-Betti-1 curves obtained from the acidification of NaCas with 1.8\% or 3.5\% GDL. In each case, the trend of the max-Betti-1 curves is similar as initially, max-Betti-1 curves have a lag phase around a Betti value of 120,000, followed by a sharp drop to a minimum (approx. 20,000). After the Betti minimum, there is an increase in Betti values.

\newcommand{\panelHplots}{5.8cm}
\newcommand{\panelHsted}{4.8cm} 

\begin{figure}[H]
\centering

\begin{subfigure}[t]{0.49\textwidth}
  \centering
  \includegraphics[width=0.92\linewidth,height=\panelHplots,keepaspectratio]{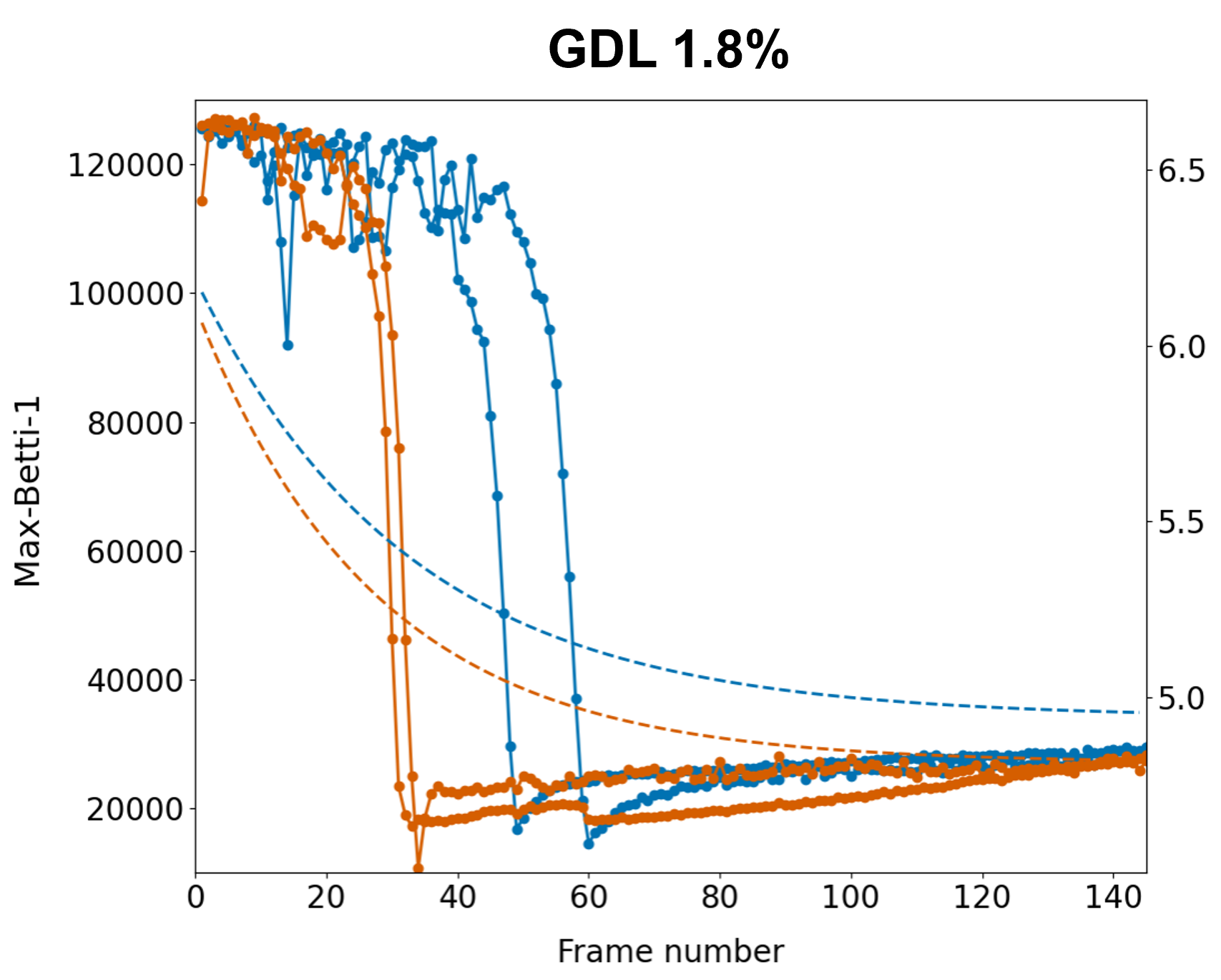}
  \caption{} \label{fig:FrameBettiph_reps_1.8}
\end{subfigure}
\hfill
\begin{subfigure}[t]{0.49\textwidth}
  \centering
  \makebox[\linewidth][l]{%
    \includegraphics[width=0.85\linewidth,height=\panelHplots,keepaspectratio]{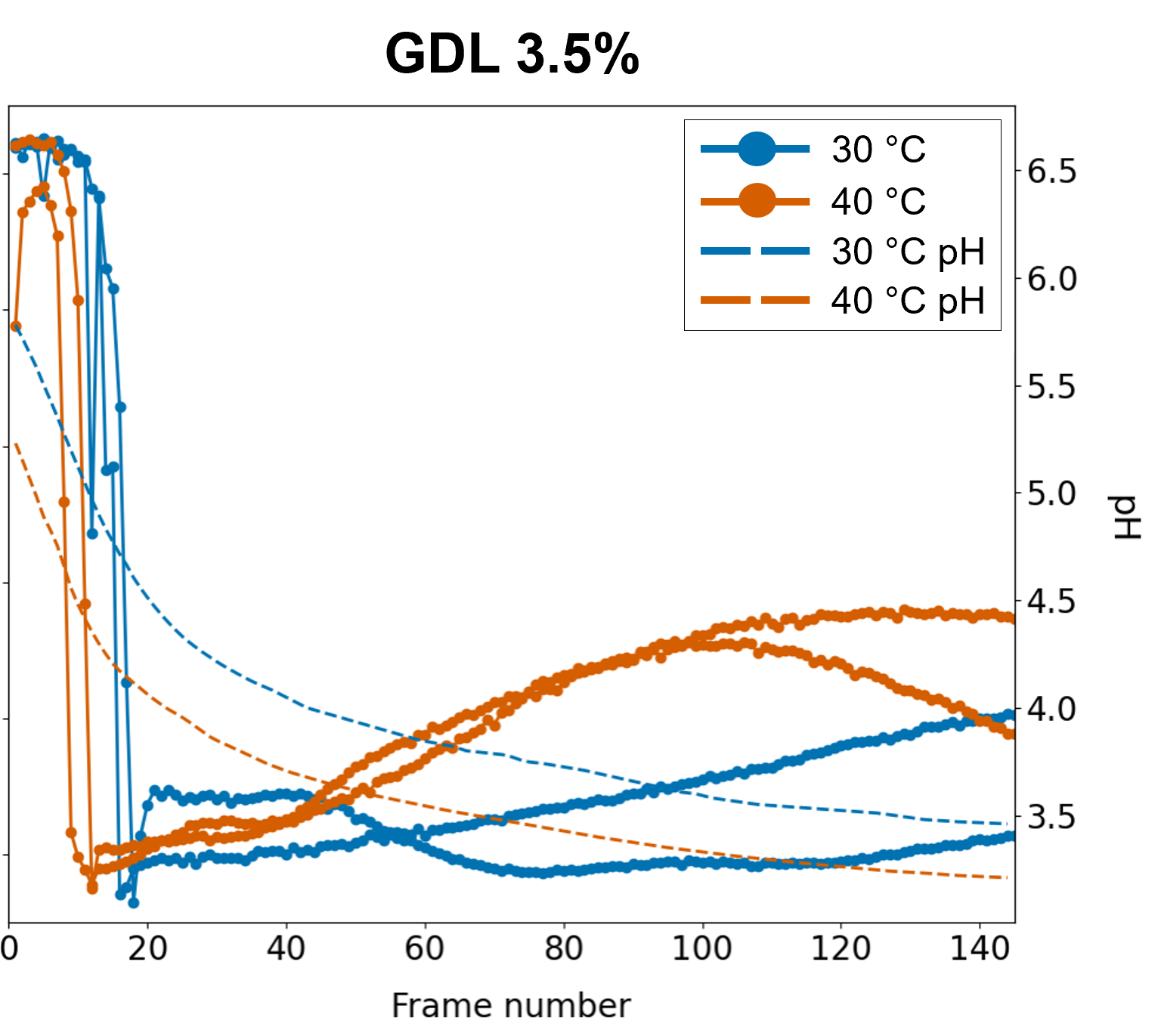}%
  }
  \caption{} \label{fig:FrameBettiph_reps_3.5}
\end{subfigure}

\caption*{\textbf{(a--b)} Topological Max-Betti-1 curves (solid lines) of the evolution of NaCas gels induced by 1.8\% (a) or 3.5\% w/v (b) glucono-$\delta$-lactone (GDL) acidification at 30 °C (blue) and 40 °C (orange). pH profiles (dashed lines) show temperature-dependent progression at 30 °C (blue) and 40 °C (orange). Same colors represent different replicates. Each frame was captured every 25 s. Note that pH axes have a different scale due to different acidification rates.}

\vspace{1.2em}

\begin{subfigure}[t]{0.49\textwidth}
  \centering
  \includegraphics[width=0.92\linewidth,height=\panelHsted,keepaspectratio]{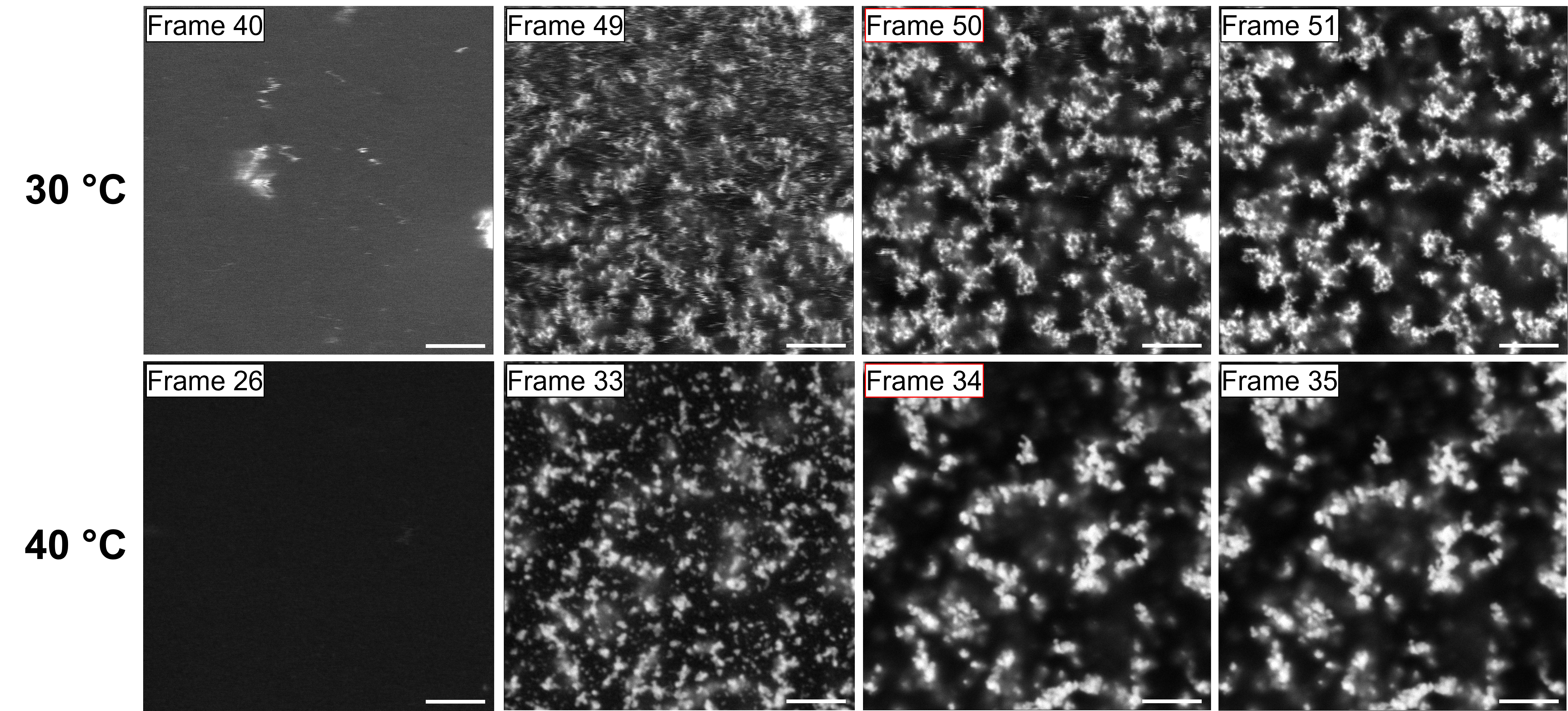}
  \caption{} 
  \label{fig:STED_1.8}
\end{subfigure}
\hfill
\begin{subfigure}[t]{0.49\textwidth}
  \centering
  \includegraphics[width=0.85\linewidth,height=\panelHsted,keepaspectratio]{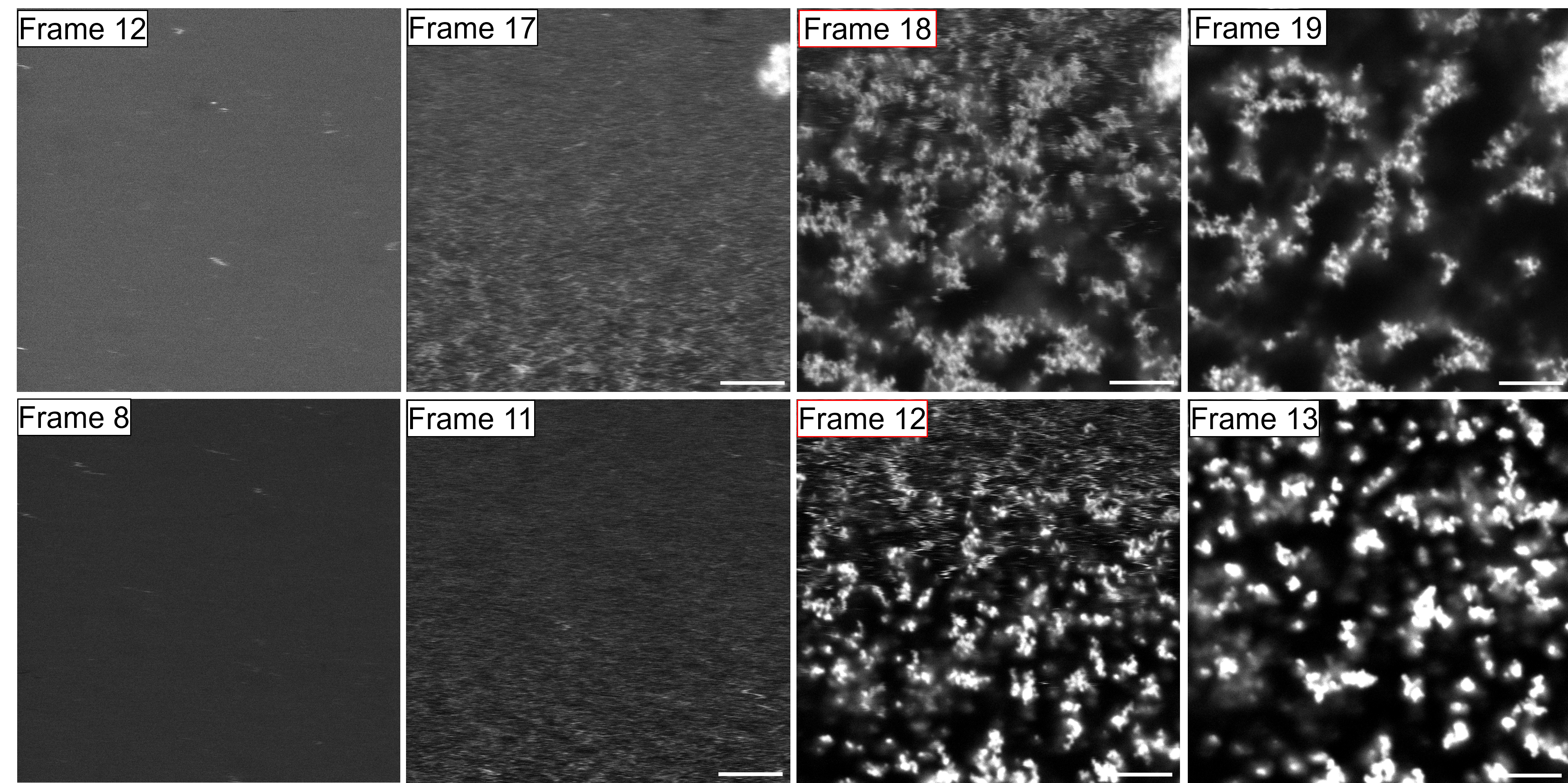}
  \caption{} \label{fig:STED_3.5}
\end{subfigure}

\caption*{\textbf{(c--d)} Representative STED micrographs of the evolution of NaCas gels induced by 1.8\% (a) or 3.5\% w/v (b) glucono-$\delta$-lactone (GDL) acidification at 30 °C (upper row) and 40 °C (lower row). Each frame was captured every 25 s. All scale bars = 5 $\mu$m.}

\caption{Topological evolution and corresponding STED micrographs of sodium caseinate gelation under different GDL concentrations and temperatures.}
\label{fig:Gelation_combined}

\end{figure}


The trends observed in the max-Betti-1 graphs are consistent with protein aggregation processes. As the pH decreases in the system due to the hydrolysis of GDL into gluconic acid, casein proteins $ (\alpha_{s1}, \alpha_{s2}, \beta, \kappa) $ aggregate into small and reversible nanoaggregates ~\cite{madadlou2010acid, braga2006effect} due to the reduction of electrostatic repulsion and hydrophobic interactions between the exposed protein hydrophobic domains ~\cite{ braga2006effect}. 

The obtained STED micrographs of the systems (Figures~\ref{fig:STED_1.8}, and ~\ref{fig:STED_3.5}) initially show from a topologically perspective large Betti values, which are associated with a disconnected heterogeneous dispersion where each casein aggregate gives rise to a close loop. As time progresses, there is a decrease in the Betii values, which topologically translates to the merging of loops. From a protein network perspective this corresponds to aggregate growth, which eventually become a network with increased connectivity.

The shorter lag phase at higher temperature and higher GDL concentration suggests an accelerated process, since increased temperatures lead to higher acidification rates and thermal motion of proteins these allow enhanced hydrophobic interactions of caseins promoting interactions leading to faster protein aggregation ~\cite{lucey1997properties}. The acidification rate also increases with a higher GDL concentration, so the lag phase is shorter for 3.5\% GDL due to fast aggregation and longer for 1.8\% as it has a more gradual casein aggregation.

Following, the drop in Betti values reaches a minimum Betti value. For GDL 1.8\% at 30 °C and 40 °C this are around frames 50 and 35, respectively (Figure~\ref{fig:FrameBettiph_reps_1.8}), while for GDL 3.5\% at 30 °C and 40 °C, the Betti minimum was around frames 18 and 12, respectively (Figure~\ref{fig:FrameBettiph_reps_3.5}). 
From the micrographs of the systems (Figures~\ref{fig:STED_1.8}, ~\ref{fig:STED_3.5}), we observed that the point where the max-Betti-1 values reach a minimum corresponds to a transition from loose small protein aggregates with apparent motion, to a percolated gel network. So, this is the first strong indication that TDA reliably captures protein aggregation dynamics, and it is possible to quantify the micro-gelation point via max-Betti-1 numbers.

To analyze the Betti curves with respect to the change in pH, the acidification profiles of the systems were also plotted in Figures~\ref{fig:FrameBettiph_reps_1.8}, \ref{fig:FrameBettiph_reps_3.5} (dashed lines). According to these, the final pH (after 1h) of the 1.8\% GDL system was 5.0 and 4.8 at 30 °C and 40 °C, respectively. For the 3.5\% GDL system, the final pH was 3.4 and 3.2 at 30 °C and 40 °C, respectively. The lower final pH for 3.5\% GDL was expected due to the high concentration, while the lower pH at higher temperature is due to the increased GDL hydrolysis at higher temperatures ~\cite{lucey1997properties}. Table \ref{Table_time_pH} summarizes the max-Betti-1 values at the start of the decay and at the minimum with respect to frame number, time (min), and pH. The start of decay was considered the point where the values deviated from the plateau linearity.

\begin{table}[H]
\centering
\caption{Betti-max1 values over time and pH during the acidic gelation of NaCas.}
\resizebox{0.8\textwidth}{!}{%
\begin{tabular}{|c|c|c|c|c|c|c|c|}
\hline
\multicolumn{2}{|c|}{} & \multicolumn{3}{c|}{\textbf{GDL 1.8\%}} & \multicolumn{3}{c|}{\textbf{GDL 3.5\%}} \\
\hline
\multicolumn{2}{|c|}{} & \textbf{Frame number} & \textbf{Time (min)} & \textbf{pH} & \textbf{Frame number} & \textbf{Time (min)} & \textbf{pH} \\
\hline
\textbf{30 °C} & {Decay of max-Betti-1 value} & \textbf{40} & \textbf{10} & \textbf{5.4} & \textbf{12} & \textbf{5} & \textbf{5.0} \\
\hline
\textbf{30 °C} & {Minimum max-Betti-1 value} & \textbf{50} & \textbf{21} & \textbf{5.2} & \textbf{18} & \textbf{7.5} & \textbf{4.6} \\
\hline
\textbf{40 °C} & {Decay of max-Betti-1 value} & \textbf{26} & \textbf{10.4} & \textbf{5.3} & \textbf{8} & \textbf{3} & \textbf{4.7} \\
\hline
\textbf{40 °C} & {Minimum max-Betti-1 value} & \textbf{34} & \textbf{14} & \textbf{5.2} & \textbf{12} & \textbf{5} & \textbf{4.4} \\
\hline
\end{tabular}%
}
\label{Table_time_pH}
\end{table}

The systems with 3.5\%GDL, have the decay of the max-Betti-1 values at pH 5.0 (30 °C) and 4.7 (40 °C), while the minimum was at a lower pH, 4.6 (30 °C) and 4.4 (40 °C). These values align with protein gelation mechanisms, where small protein aggregates start to form at reduced pH of approx. 5.0 ~\cite{madadlou2010acid, braga2006effect, dalgleish2005mechanism}, while near isoelectric point of caseinate at 4.5, the net negative charge is neutralized leading to non-covalent aggregation, higher attractive hydrophobic and van der Waals interactions, irreversible protein aggregation and consequently a percolated casein network ~\cite{lucey1997properties, raak2017enzymatic}. 
The slightly lower pH values at 40 °C for both decay and minimum Betti numbers correspond to the higher thermal motion and higher hydrolysis rate of GDL at higher temperatures that increase protein-protein interactions giving a faster network formation ~\cite{lucey1997properties}. At lower GDL (1.8\%), the pH values of both the decay and minimum are higher than in the case of 3.5\%GDL. This is related to the slower acidification rate at this GDL concentration as it allows the system to have more time at a higher pH 5.5–5.0 (Figure~\ref{fig:FrameBettiph_reps_1.8} and~\ref{fig:FrameBettiph_reps_3.5}, dashed lines) where attractive interactions drive aggregation, thus leading to the formation of a gel at a higher pH.
The correlation between Betti decay and minimum values with the transitions observed in the micrographs and the pH-based mechanisms of protein aggregation corroborates that TDA can accurately quantify the dynamics of casein network percolation. Given that the quantification captures slight differences between the systems, TDA appears to be a highly sensitive method for detecting structural changes.

\begin{figure*}[!t]

\centerline{\includegraphics[width=1.0\textwidth]{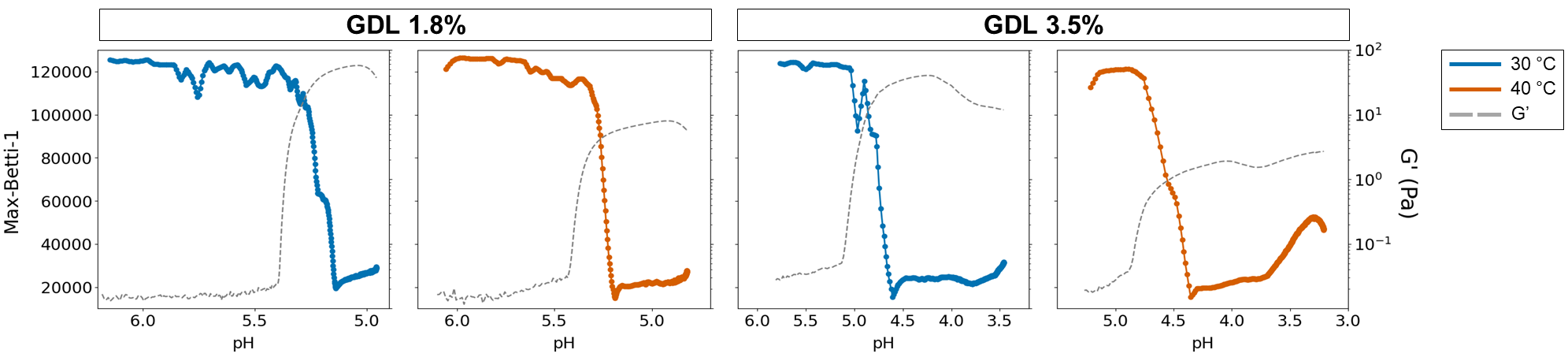}}
\caption{Topological max-Betti-1 curves (solid lines) and rheological parameter storage modulus (G’) (dashed lines) over pH at 30 °C (blue) and 40 °C (orange), with GDL concentration of 1.8\% or 3.5\%.}
\label{fig:SAOS}

\end{figure*}

Small amplitude oscillatory rheology was used to relate max-Betti-1 values to the mechanical response of network development. Results are depicted in Figure~\ref{fig:SAOS}, where max-Betti-1 values are plotted over pH together with G’ at both GDL concentrations, 1.8\% or 3.5\%. 

In all cases, there is a sharp increase in the storage modulus (G'), which from a rheological perspective is related to the transition from a weakly connected, energy dissipative structure to a solid-like elastic network, characteristic of the sol-gel transition ~\cite{lucey1997properties}. This increase in G' closely aligns with the decay of max-Betti-1 values, described above as the progressive merging of loop-like structures as the networks densify during acidification. So, this suggests that the bulk elastic network formation aligns with the microstructure undergoing topological organization. And, since the max-Betti-1 minimum coincides with the onset of the G' plateau, it marks the transition to a maximum topological simplification. After this minimum, the network continues to evolve, which is discussed below.

The correlation of the max-Betti-1 curves with the micrographs (Figures~\ref{fig:STED_1.8}, ~\ref{fig:STED_3.5}), the acidification mechanism changes (Table \ref{Table_time_pH}), and the mechanical response of the gel formation (Figure~\ref{fig:SAOS}) confirms that max-Betti-1 decay and minimum values correspond to the formation of protein aggregates with a liquid-like behavior as the network percolates and becomes highly elastic due to the decrease in pH. The pH values where the viscoelastic transition is observed (Figure~\ref{fig:SAOS}) agree with previous reports of acid-induced casein gel mechanisms \cite{lucey1997properties, braga2006effect, raak2017enzymatic}. 

These results are consistent with a previous particle tracking microrheology report of a similar system \cite{moschakis2010kinetics}. Moschakis et al. \cite{moschakis2010kinetics} showed that probe particles had a high mobility during the initial acidification phase, similar to the observed high max-Betti-1 values. Probe particles were eventually dynamically arrested as the gel network was formed. Similarly to the topological percolation that was observed (decrease-minimum in max-Betti-1 values), this drop of mobility was correlated with the gel point, where G' increases sharply. These results demonstrate that the arrest in particle movement reported by Moschakis et al. \cite{moschakis2010kinetics} is indeed due to a network percolation. 

\begin{figure}[H]
\centering

\begin{subfigure}{0.8\textwidth}
\centering
\includegraphics[width=0.75\linewidth]{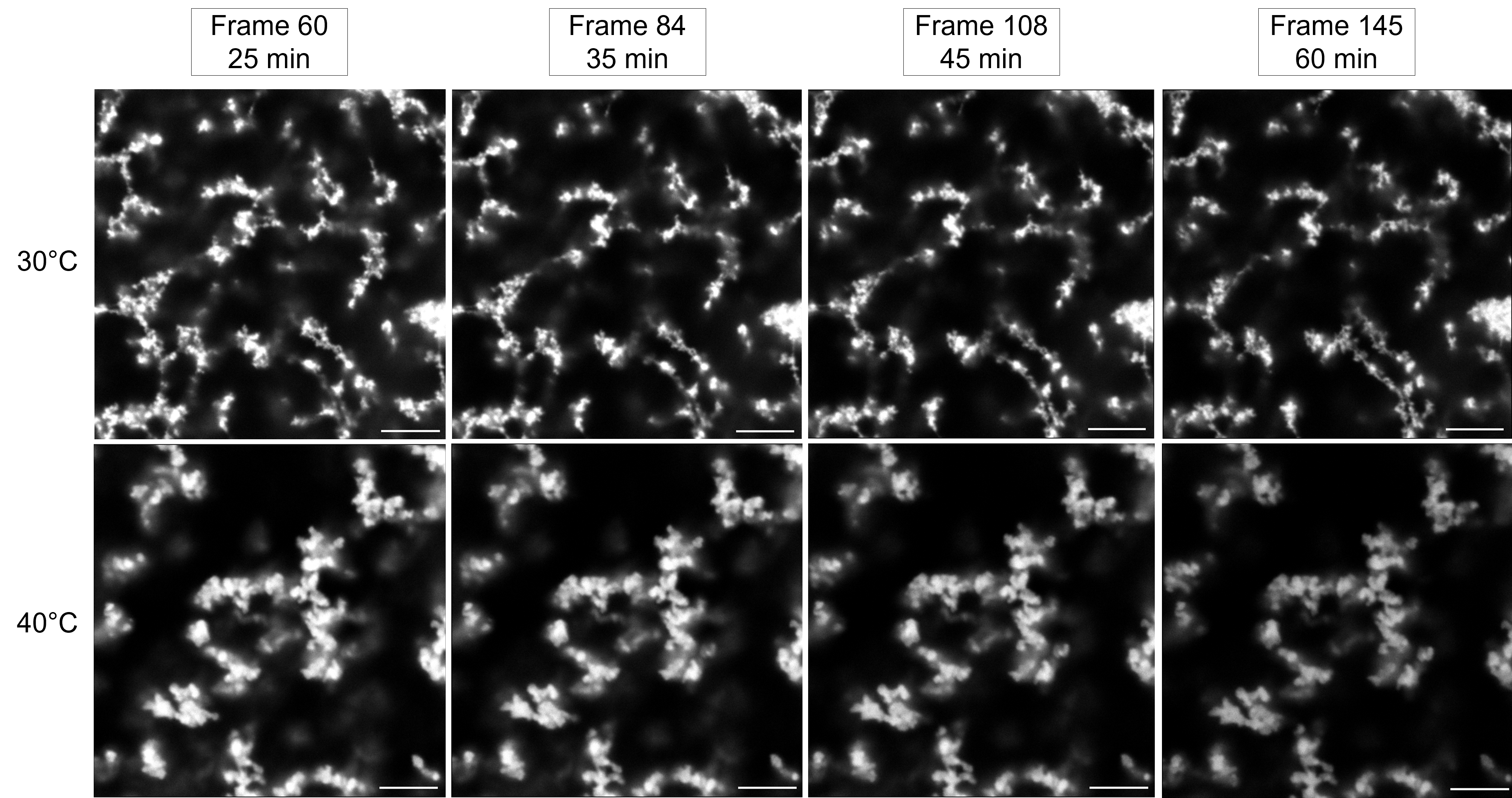}
\caption{}
\label{fig:STED_1.8_Phase2}
\end{subfigure}

\vspace{0.8cm}

\begin{subfigure}{0.8\textwidth}
\centering
\includegraphics[width=0.75\linewidth]{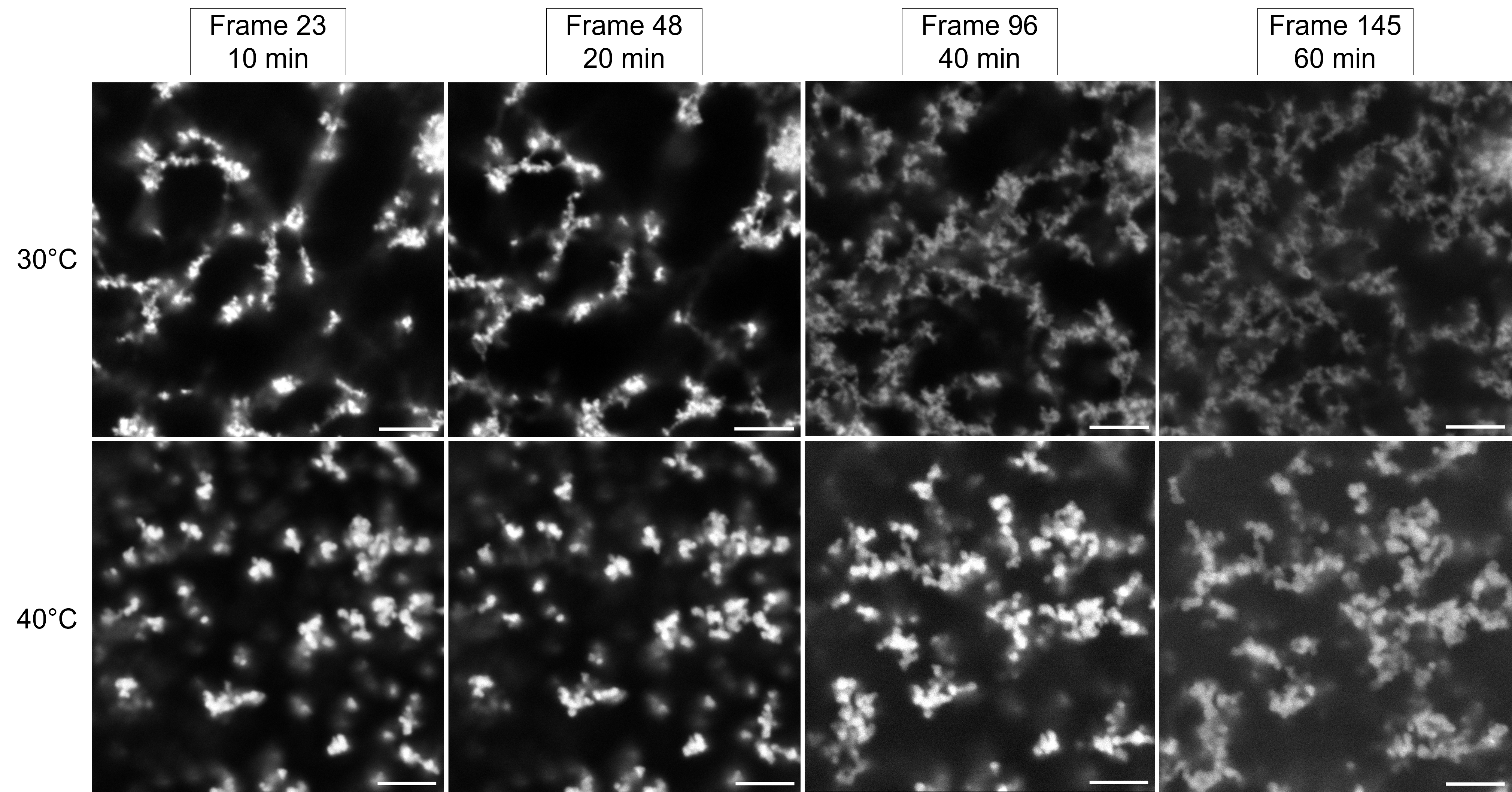}
\caption{}
\label{fig:STED_3.5_Phase2}
\end{subfigure}

\caption{STED micrographs of the evolution of NaCas gels in the rearrangement phase, after the gel point and until the end of aggregation (1h). System gelation induced by either 1.8\% (a) or 3.5\% (b) glucono-$\delta$-lactone (GDL) with acidification at 30 °C (first row) and 40 °C (second row). Since each frame was captured every 25 s, the time in seconds is noted in each micrograph. All scale bars= 5 $\mu$m.}
\label{fig:STED_Phase2_combined}

\end{figure}

After the Betti minimum, there is a constant increase in Betti values (Figure~\ref{fig:SAOS}). Overall, such increase is evident for 3.5\% GDL at 40 °C, followed by 30 °C, and minimum increase for the system with 1.8\% at both temperatures. In terms of protein aggregation, once the protein network is established, a rearrangement phase occurs \cite{chen2000temperature, lucey1998comparison}. This phase involves structural reorganization of the network due to continued crosslinking, strengthening of casein strands, and local particle rearrangement. Topologically, this structural rearrangement was quantified as an increase in loops suggesting that the network was more disconnected. By looking at the system with 3.5\% GDL at 40 °C, which has a sharper increase in Betti values, the evolution of the network to a more disconnected one is consistent with the higher acidification rate that causes enhanced hydrophobic interactions and rearrangement processes \cite{lucey1997properties}. The topologically more open network in this system, in comparison to the other systems, can be observed in the representative micrographs of the rearrangement phase (Figures~\ref{fig:STED_1.8_Phase2}, \ref{fig:STED_3.5_Phase2}). The lower increase in Betti values in the other systems, at 30 °C in both cases and at 40°C with 1.8\%GDL, corresponds to the lower acidification rate that allowed the gel to form gradually, and which is known to lead to finer more stable strand-like chains with a relatively homogeneous interconnected network and minimal post-percolation rearrangement ~\cite{lucey1998comparison, braga2006effect}.

Correlating these observations in the rearrangement phase with rheological parameters, the higher values of Betti in the 3.5\% systems at pH around 3.5 correlate with a decrease in the gel strength (G’). Raak et al. \cite{raak2017enzymatic} previously linked this rheological profile to reduced protein-protein interactions caused by increased electrostatic repulsion below the isoelectric point. Again, TDA quantified this as an increase in loop count due to the more disconnected network. So the increase in the TDA curves at this pH can indeed be linked to protein electrostatic interactions and thus a transition to a more disconnected network. 

Previously, Smith et al. \cite{smith2024topological} used TDA to quantify topological deformations of soft gels through computer simulations and oscillatory shear. Results showed that topological transitions of the gel under shear could also be directly correlated with physical phenomena such as shear stiffening or hardening \cite{smith2024topological}. Thus, our framework elaborates on what Smith et al. reported with simulations, but also shows that TDA is able to quantify microstructural reorganization with a sensitivity and spatial resolution that complement the characterization of subtle averaged changes in bulk rheology observed by Raak et al. and others.

In summary, Betti curves effectively captured the topological dynamic evolution of acid casein gelation, which also correlated with an established rheological marker, the storage modulus. At the beginning, the protein aggregates were resolved by TDA as a high number of loops. As the acidification of the system progressed, max-Betti-1 curves showed a decrease related to the closure of TDA loops, which we connect to the higher aggregation of caseins and eventual emergence of a topologically spanning network. This percolation point was in line with the initial elastic response, the gel point. Post-percolation, in the rearrangement phase, TDA tracked subtle structural transitions of the network as protein particles further interact, which also correlated with the viscoelastic response of the gel.


Moving beyond topological descriptors, we next quantified network irregularity using fractal-based methods. The Differential Box-Counting (DBC) method measures structural complexity, so as aggregation and rearrangement progress, we expect DBC to decrease over time, indicating a transition toward a more stable and regular pattern. In parallel, the MultiFractal Partition (MFP) captures the heterogeneity of scaling behavior within the structure, where it is expected an initial high fractality due to the greater heterogeneity, followed by a stabilization phase as the network consolidates.

\begin{figure*}[!htbp]
\begin{center}
\centerline{\includegraphics[width=0.80\textwidth]{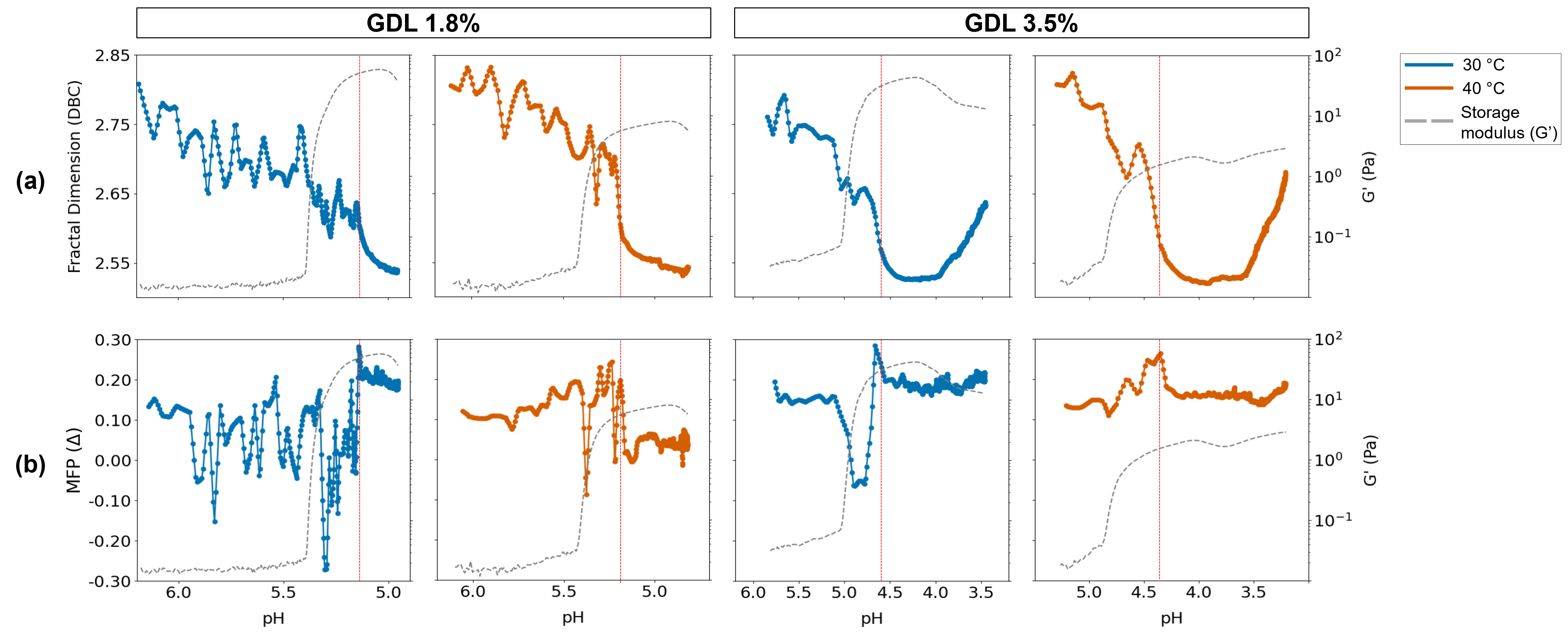}}
\caption{Fractal dimension via either DBC method (a) or via MFP method (b) over pH, at 30 °C (blue) and 40 °C (orange), with GDL concentration of 1.8\% or 3.5\%. Red vertical line marks the percolation point obtained via max-Betti-1 minimum.}
\label{fig:FD}
\vspace{-0.9cm}
\end{center} 
\end{figure*}

Figure \ref{fig:FD}a shows the evolution of the fractal dimension (FD) measured via the Differential Box Counting (DBC) method as a function of pH. Replicate experiments showed similar trends across both GDL concentrations and temperatures, confirming the reproducibility of the method, thus all graphs represent the average of such replicates. 
Across all systems, the FD begins high and decreases progressively over acidification before reaching a plateau. In the systems with higher GDL concentration, a secondary increase in FD is observed at low pH. This behavior tracks the structural evolution of the gel as observed with STED (Figures~\ref{fig:STED_1.8}, ~\ref{fig:STED_3.5}). The initial high FD indicates a space-filling, image-based texturally homogeneous dispersion of protein aggregates. As pH decreases, the protein aggregates grow and form an open, ramified network structure. This structural transition is quantified by DBC as a reduced topographic complexity, since a spatially organized network produces a more regular intensity surface than a uniform dispersion, and thus a lower fractal dimension. The plateau marks the stabilization of the network connectivity, consistent with the general observation that fractal dimension decreases during the early stages of aggregation as structural ordering increases \cite{chakraborti2003changes}. 

Previous studies on acid-induced sodium caseinate gels report final fractal dimensions characteristic of a Reaction-Limited Cluster Aggregation (RLCA) regime \cite{bremer1990fractal, bremer1993formation},  \cite{vetier1997effect}. Direct numerical comparisons with literature values are not straightforward, however, as traditional approaches rely on binary box-counting or scattering models that characterize the 2D or 3D geometric shape of clusters \cite{bremer1990fractal, bremer1993formation, vetier1997effect, chardot2002growth}. In contrast, our DBC method measures the topographic complexity of grayscale intensity surfaces, a texture-based metric with a different theoretical range \cite{sarkar1994efficient} that systematically yields higher absolute values than binary projection methods. As our approach captures intermediate stages of active aggregation over the first hour of acidification i.e. dynamically evolving structures, comparisons with existing research that typically characterize the final, static gel structure, is limited. 

Dynamic tracking of fractal dimension during acid-induced casein aggregation has been reported using light scattering \cite{chardot2002growth}, corroborating that FD is inherently dynamic rather than fixed. Critically, while both approaches reach the conclusion that fractal dimension is not a static constant but a dynamic parameter that tracks the specific stages of protein network formation, our combination of STED super resolution microscopy and DBC reveals the earliest aggregation dynamics, a regime inaccessible to light scattering due to the size-dependent resolution limit. 

The red vertical line in Figure \ref{fig:FD}a marks the point where the minimum in the max-Betti-1 curves was observed, denoted as the max-Betti-percolation point. This coincides with the point where FD deviates from a linear decay in all systems, an indication that the DBC values are structurally consistent with the topological percolation of the network, and suggests that it can serve as a complementary indicator of gelation onset. 

For the systems with 3.5\% GDL, the DBC method reveals after the percolation point a plateau phase followed by a further increase in FD. The plateau reflects network stabilization once sufficient connectivity is established, while the further FD increase represents post-percolation structural aging. Continued compaction of fractal flocs, cluster growth, and strand coarsening increase the mass density of the network strands \cite{cipelletti2000universal}, producing stronger contrast between protein-dense and void regions and thus higher topographic complexity as captured by DBC. The secondary rise is attributed to the faster acidification kinetics with higher GDL concentration or temperature, which promote more extensive network reorganization post-percolation \cite{lucey1997properties}.

In contrast, the systems with 1.8\% GDL show a plateau without a pronounced secondary rise, consistent with reduced particle rearrangement, fewer particle-particle contacts, and limited strand coarsening at the slower acidification rate.


Figure \ref{fig:FD}b shows the multifractal characteristics of the networks via the partition width of the multifractal spectrum. In all systems, the width is initially high and then decreases as it stabilizes, showing the rapid transition from a disordered, high-entropy protein dispersion to an organized network. The point of stabilization aligns with the topological minimum in the TDA curves (red vertical line) and the FD transition discussed above, which suggests that MFP can also act as an independent indicator of gel percolation.

The stabilization of the MFP values suggests that the network densification after percolation follows spatially homogeneous dynamics, rather than generating localized heterogeneities such as dense clusters and sparse voids. The stability of the width could also reflect competing structural effects that offset each other in the partition function, or a reduced sensitivity of the binary-based MFP descriptor to the type of intensity reorganization that DBC captures in the grayscale domain. However, the homogeneous coarsening behavior interpretation is consistent with the aging dynamics of colloidal gels \cite{cipelletti2000universal} and remains the most coherent considering the presented structural evidence.

\begin{figure}[H]
\begin{center}
\centerline{\includegraphics[width=0.80\textwidth]{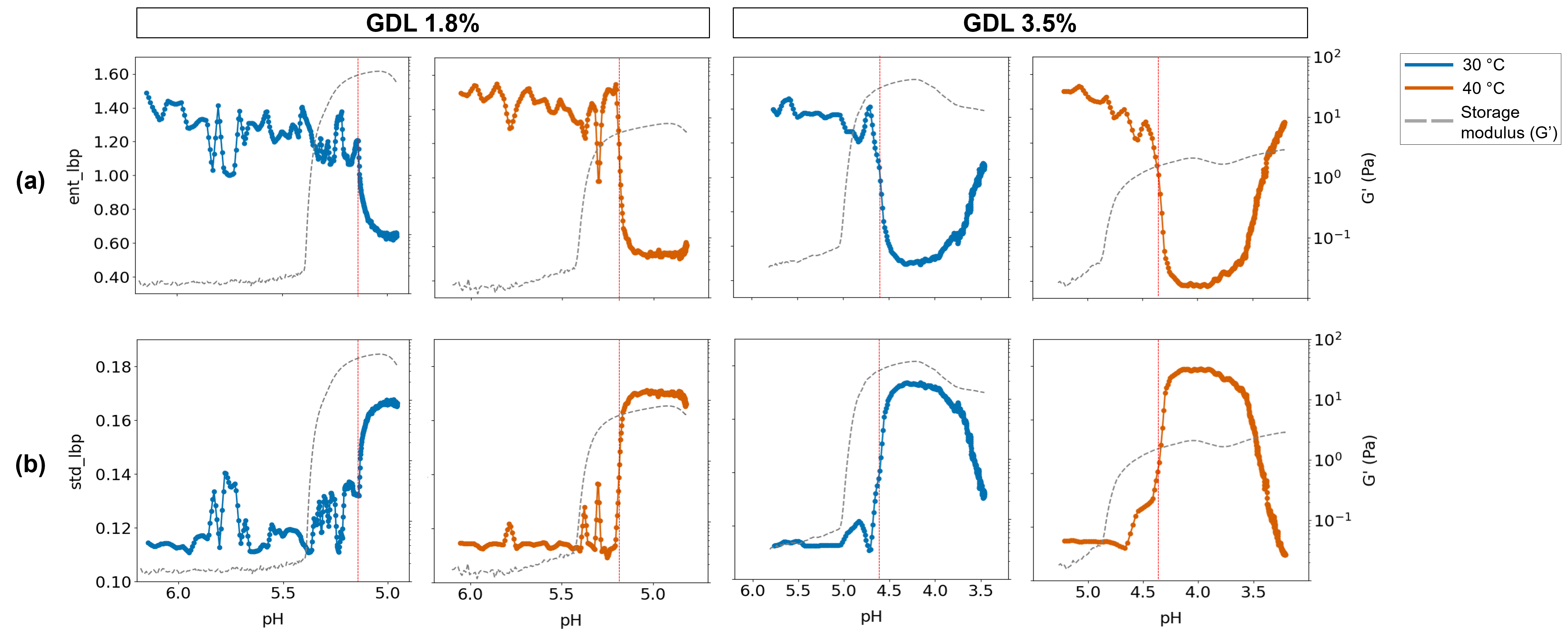}}
\caption{Entropy (a) and standard deviation (b) via the local binary pattern method, and storage modulus (G') over pH, at 30 °C (blue) and 40 °C (orange), with GDL concentration of 1.8\% or 3.5\%. Red vertical line marks the percolation point obtained via max-Betti-1 minimum.}
\label{fig:LBP}
\vspace{-0.9cm}
\end{center} 
\end{figure}

As a final step, we applied Local Binary Pattern (LBP) to the dynamic set as a powerful texture descriptor that labels pixels by comparing them to their neighbors. In this study, we considered this method to analyze the complexity of the structure over time. Figure~\ref{fig:LBP} shows the changes of the texture LBP patterns illustrated by the entropy (a) and standard deviation (std) (b) for the systems with 1.8\% GDL or 3.5\% GDL at 30 °C (blue) and 40°C (orange), together with the mechanical response, storage modulus (G'). Again, the red line denotes the sol-gel transition observed via the max-Betti-1 transition, while the graphs correspond to the average of two replicates.

According to Figure~\ref{fig:LBP}, the distribution of the texture patterns (Figure~\ref{fig:LBP}a) and the dominant texture (Figure~\ref{fig:LBP}b) can be followed as they correlate with the observations in STED micrographs (Figures~\ref{fig:STED_1.8}, ~\ref{fig:STED_3.5}). At the beginning, the entropy curves with 1.8\%GDL show an initial oscillatory period, fluctuating between approximately 1.5 and 1.0, while with 3.5\% this fluctuation is less, between 1.5 and 1.2. Since after this period there is a transition to a lower entropy, this marks a point where the texture of the structure stabilizes around a subset of dominant structural patterns, which can be considered the gel percolation. This is corroborated by the alignment of this transition point to the point where the max-Betti-1 values reached a minimum, similarly considered the gel point, and where fractal characteristics showed a transition to an organized structure.

Fluctuations at the start of aggregation can also be observed in the case of standard deviation curves, which are also greater with 1.8\% GDL. In this case, std curves show an increasing trend since structural patterns begin to dominate. Since this change occurs at the same point where entropy decreases, it is also considered a structure consolidation.

While bulk rheology shows a low gel strength at the beginning with a sudden increase as the network forms, LBP quantification is able to resolve the textural transitions underlying such rheological changes at the microscopic level.

Differences of the initial fluctuation period of entropy and std curves between 3.5 and 1.8\% GDL correspond with their different acidification rates. The slower rate with 1.8\% allows protein particles to have transient microstructural changes and reorganization for an extended time before the network formation. Overall, the LBP transitions agree with the mechanical response of gelation, which demonstrates that this quantification technique is sensitive to differences in acidification kinetics and the resulting gelation dynamics.

After the stabilization of the network, the curves show a plateau phase with less noticeable changes and, for the system with 3.5\% GDL, a further increase in entropy and decrease in std (Figure~\ref{fig:LBP}). This behavior indicates that LBP can also track subtle changes in the rearrangement phase of the network, which again, correspond to the decrease in pH and consequent protein-protein interactions or due to temperature-driven enhanced protein interactions \cite{lucey1997properties}. This pattern corresponds with a shoulder in G' reflecting a more energy dissipative network, but LBP shows that this is due to the broadening of the local texture pattern distribution, meaning that the textural regularity established at the percolation point is being lost, rather than a uniform increase in network elasticity as suggested by rheology. While the progressive reorganization of the network strands and interparticle contacts in the rearrangement phase is well established \cite{chen2000temperature, lucey1998comparison, braga2006effect} characterizing these dynamics quantitatively at the microstructural level remains challenging. With the LBP analysis, a continuous quantitative descriptor of the local texture evolution is provided. 

Taken together, the four descriptors provide complementary perspectives on the structural evolution of the system. Although each method emphasizes different aspects of the system, consistent transitions are observed across descriptors, particularly around the onset of network formation. TDA captures changes in network topology, particularly the formation and reorganization of connectivity and loops. DBC reflects variations in grayscale structural complexity, while MFP quantifies multi-scale heterogeneity and the distribution of dense and sparse regions. LBP, in turn, captures local texture variations associated with fine-scale structural rearrangements. This agreement supports the robustness of the proposed framework in capturing both global and local structural changes during protein aggregation.

\section*{Limitations and future work}

The proposed framework in this paper was evaluated using both simulated fractal structures and experimental sodium caseinate gelation. The obtained results and their validation represent a proof-of-concept within a specific protein aggregation system. The selected model system provides well-characterized structural transitions that allow comparison between topological descriptors and rheological measurements. Nevertheless, the general applicability of the toolbox to other systems, such as different protein networks, colloidal gels, or polymeric structures, should be further investigated in future studies. Additional validation across diverse imaging modalities and microstructural processes would help establish the broader robustness of the proposed approach.

\section*{Conclusion}
In this paper, we propose the combination of super-resolution microscopy with an analytical toolbox that integrates Topological Data Analysis (TDA), Differential Box Counting (DBC), Multifractal Partition (MFP), and Local Binary Patterns (LBP) to capture the dynamics of casein protein aggregation from local texture evolution to global topological transitions. With STED microscopy, we tracked the microstructural evolution of the gelation. With the use of the cubical complex technique in TDA, we applied several filtration values to the images based on each unique pixel value. This saved the information and removed the noise so all the informative topological features of the image could be quantified, without problems observed in intensity-based methods. TDA showed that gelation dynamics can be tracked via topological loops up to the sol-gel transition and further into the network rearrangement phase. In all cases these transitions were correlated with the mechanical response of the network over time. Fractal methods captured the dynamics of fractal dimension over time via DBC, and the transition from high-entropy dispersion to organized structures via MFP, while LBP metrics highlighted the emergence of dominant local textures. 
This integrated analysis captured subtle aggregation dynamics in the rearrangement phase, linked with averaged-metrics in traditional bulk rheology, offering an enhanced quantitative understanding of gelation mechanisms post-percolation. As we aim to understand food-structure relationships, we envision our methodology can be used in other systems with minimal adjustments. All codes are available at \href{https://github.com/Zahratabatabaei/Delifood_CV_paper.git}{this GitHub repository}.

\section*{Acknowledgement}

Image acquisition was performed at the Danish Molecular Biomedical Imaging Center (DaMBIC, University of Southern Denmark), supported by the Novo Nordisk Foundation (NNF) (grant agreement number NNF18SA0032928).

\section*{Credit authorship contribution statement}

Zahra Tabatabaei: Experiments and programming, Writing – original draft, Writing – review \& editing, Visualization, Investigation, Methodology, Formal analysis, Data curation, Conceptualization. 

Diana Soto-Aguilar: Data collection, Writing – original draft, Writing – review \& editing, Visualization, Investigation, Methodology, Formal analysis, Data curation, Conceptualization.

José C. Bonilla: Funding acquisition, Writing – review, Methodology, Supervision, Conceptualization.

Mathias P. Clausen: Funding acquisition, Project administration, Writing – review, Methodology, Supervision, Conceptualization.

Jon Sporring: Funding acquisition, Project administration, Writing – review, Methodology, Supervision, Conceptualization. 
\section*{Data availability}
The datasets and code generated and analyzed during the current study are available at: 
\hyperlink{https://github.com/Zahratabatabaei/Delifood_CV_paper}{Github Link}

\section*{Funding}
The work is presented in this paper is funded by the Villum Fonden grant number 00057398.

\section*{Competing interests}
The authors declare no competing interests.
\bibliography{sample}

\newpage
\section*{Supplementary Data}
\setcounter{figure}{0} 
\renewcommand{\thefigure}{S\arabic{figure}} 

\begin{figure*}[!htbp]
\begin{center}
\centerline{\includegraphics[width=0.60\textwidth]{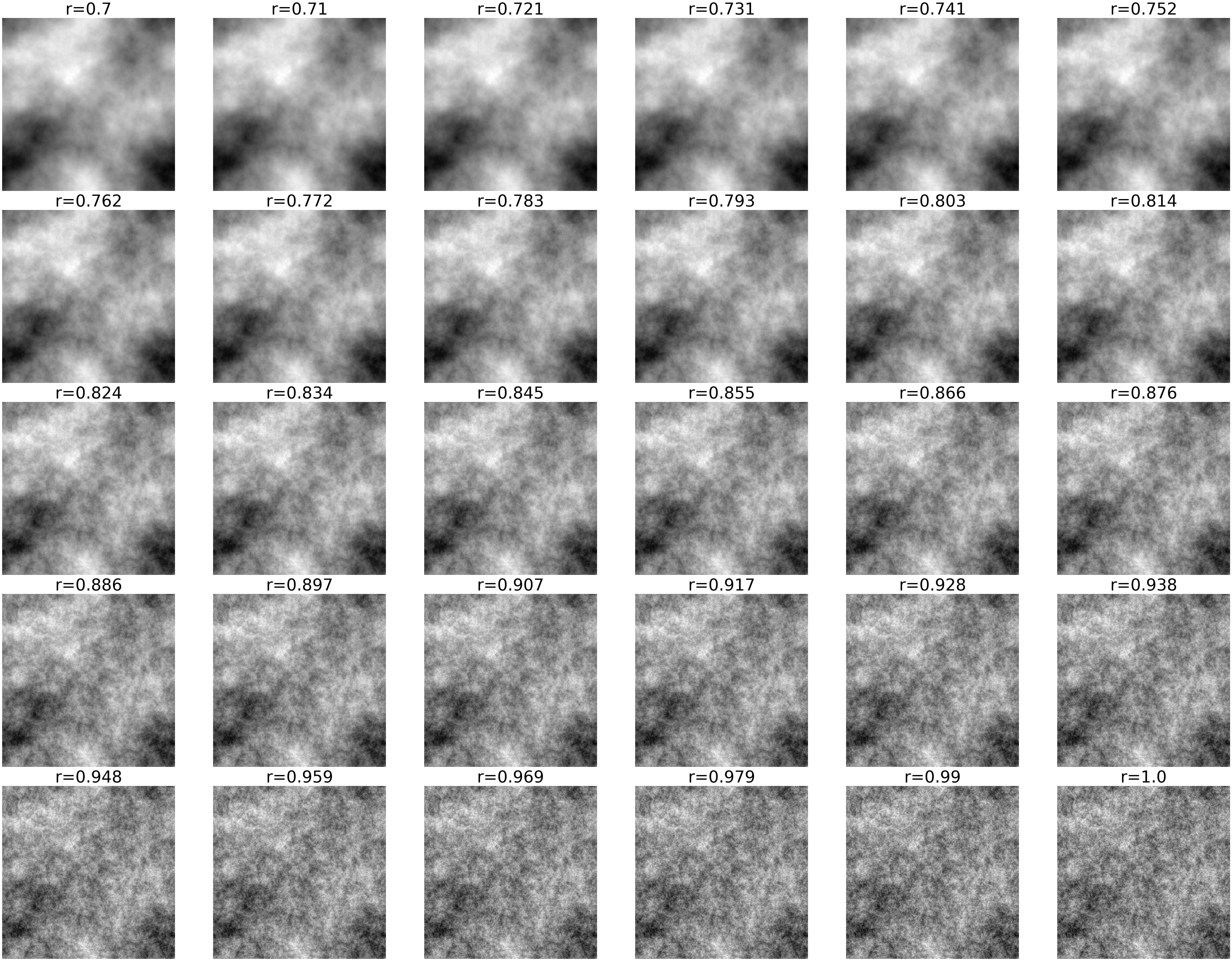}}
\caption{Simulated set of the fractal images by increasing the roughness from 0.7 to 1.}
\label{fig:simulated}
\vspace{-0.9cm}
\end{center} 
\end{figure*}


\end{document}